\pdfoutput=1

\documentclass[11pt]{article}
\usepackage{multirow}
\usepackage{graphicx}
\usepackage{amsmath} 
\usepackage[final]{acl}

\usepackage{times}
\usepackage{latexsym}

\usepackage[T1]{fontenc}

\usepackage[utf8]{inputenc}

\usepackage{microtype}

\usepackage{inconsolata}

\title{Unveiling Selection Biases: Exploring Order and Token Sensitivity in Large Language Models}

\author{
    Sheng-Lun Wei\textsuperscript{1} \quad
    Cheng-Kuang Wu\textsuperscript{1} \quad
    Hen-Hsen Huang\textsuperscript{2}\quad
    Hsin-Hsi Chen\textsuperscript{1}
    \\
    \textsuperscript{1}National Taiwan University, Taiwan
    \\
    \textsuperscript{2}Academia Sinica, Taiwan
    \\
    \texttt{\{weisl,ckwu\}@nlg.csie.ntu.edu.tw}
    \\
    \texttt{hhhuang@iis.sinica.edu.tw\quad hhchen@ntu.edu.tw}
}

\begin{document}
\maketitle
\begin{abstract}
In this paper, we investigate the phenomena of "selection biases" in Large Language Models (LLMs), focusing on problems where models are tasked with choosing the optimal option from an ordered sequence.
We delve into biases related to option order and token usage, which significantly impact LLMs' decision-making processes.
We also quantify the impact of these biases through an extensive empirical analysis across multiple models and tasks.
Furthermore, we propose mitigation strategies to enhance model performance.
Our key contributions are threefold: 1) Precisely quantifying the influence of option order and token on LLMs, 2) Developing strategies to mitigate the impact of token and order sensitivity to enhance robustness, and 3) Offering a detailed analysis of sensitivity across models and tasks, which informs the creation of more stable and reliable LLM applications for selection problems.
\end{abstract}

\section{Introduction}
Large Language Models (LLMs) have demonstrated remarkable abilities across various tasks \cite{Achiam2023GPT4TR, Anil2023GeminiAF, touvron2023llama}, leading to their widespread adoption in downstream applications.
In particular, the utilization of zero-shot or few-shot prompting techniques emerged as a highly convenient approach in harnessing the potential of LLMs, since these techniques empower end-users to solve a wide range of tasks without the need for extensive fine-tuning.

Despite LLMs' impressive performance and convenience, empirical investigations have found that their output is highly sensitive to the choice of prompts, and even subtle modifications of instructions or demonstrations have considerable influence on their performance.
To address this issue, several works have been dedicated to the identification and mitigation the inherent biases in LLMs, aiming to enhance their robustness and reliability \cite{Zhao2021CalibrateBU, Si2023MeasuringIB, Fei2023MitigatingLB}.

In this study, our focus is directed towards the domain of ``selection problem'', where LLMs are instructed to select the optimal choice from an ordered sequence of choices.
This problem encompasses a variety of downstream applications, including but not limited to classification, multiple-choice questions~\cite{Zheng2023LargeLM, Pezeshkpour2023LargeLM}, and model evaluation scenarios~\cite{Zheng2023JudgingLW, Wang2023LargeLM, Shen2023LargeLM}.
In our analysis, we identifies specific biases within the context of selection problems, which we call ``selection biases'' to encapsulate these discernible tendencies.
These biases manifest as systematic deviations in LLMs' preferences, and a thorough understanding of these biases is pivotal for enhancing the robustness of LLMs across the spectrum of applications under the scope of selection problems.
Our subsequent exploration delves into the characterization, quantification, and mitigation strategies to address these biases.
It is crucial to highlight that our analysis centers on the zero-shot setting.
This choice distinguishes our work from previous endeavors, which predominantly concentrate on few-shot settings, making it difficult to disentangle biases stemming from in-context demonstrations.

Our contributions can be summarized as follows: \textbf{1)} We quantify the influence of option order and token on the decision-making processes of various LLMs when tackling selection problems, providing clear insights into how these factors affect model performance; \textbf{2)} We introduce strategies to mitigate the effects of token and order sensitivity, leading to performance improvements across a broad spectrum of tasks; \textbf{3)} We offer a thorough understanding of the sensitivity landscape through an empirical study encompassing different models, tasks, and sensitivity settings. The analysis enables us to identify the most effective strategies for addressing sensitivity issues in diverse task scenarios.

\section{Related Work}

\textbf{Biases of LLMs.}
Several studies have delved into the biases of LLMs.
\citet{Zhao2021CalibrateBU} identifies three notable biases:
majority label bias, where LLMs exhibit a propensity to output the most frequent label in few-shot demonstrations;
recency bias, which is the tendency to repeat the label appearing towards the end of the prompt;
and common token bias, manifesting as the inclination to output tokens prevalent in the pre-training distribution.
\citet{Fei2023MitigatingLB} further identifies the domain-label bias, which could be detected and estimated using random in-domain words from the task corpus.
Additionally, \citet{Si2023MeasuringIB} focuses on feature bias, which is the tendency to use one feature over another to predict the label, even when both features in the prompt are equally effective for predicting the label.
However, these works mainly focus on the few-shot settings, which fails to disentangle the effects of selection biases from biases caused by in-context examples.

\noindent \textbf{Selection Problem of LLMs.} 
Previous studies have explored the use of LLMs in tackling selection problems. In Multiple Choice Questions (MCQs), \citet{Robinson2022LeveragingLL} demonstrated the application of LLMs to MCQs, focusing on how different prompting techniques influence the model's decision-making process. \citet{Pezeshkpour2023LargeLM} highlighted how LLMs are affected by position bias when addressing MCQs, while \citet{Zheng2023LargeLM} pinpointed token bias as the primary reason LLMs are not robust selectors in this context.
In evaluation scenarios, \citet{Shen2023LargeLM} employed LLMs to assess the abstractive summarization outcomes of models, introducing three distinct settings: reason-then-score (RTS), MCQ scores, and head-to-head comparison (H2H). \citet{Zheng2023JudgingLW} applied LLMs as evaluators in chatbot interactions, employing a two-round approach rather than a single-step evaluation. \citet{Wang2023LargeLM} discovered that LLMs' evaluation fairness is significantly compromised by option position bias, indicating that LLMs can be heavily influenced by the positioning of options.

\section{Experimental Setup}

\subsection{Evaluation Tasks}
We experiment on six multi-choice tasks with the number of choice options varying from two to five. The six benchmarks are: ARC-Challenge \cite{clark2018think}, HellaSwag \cite{zellers2019hellaswag}, MMLU \cite{hendrycks2021measuring}, Winogrande \cite{WINOGRANDE}, MathQA \cite{Amini2019MathQATI}, and OpenBookQA \cite{Mihaylov2018CanAS}.

We select these datasets due to their coverage of a wide range of domains, including commonsense reasoning, STEM, social sciences, humanities, etc. This diversity ensures a comprehensive evaluation across various fields of knowledge. Data statistics details are shown in Table \ref{tab:benchmark} in Appendix \ref{data-details-appendix} due to space constrains.

\subsection{Models}
We adopt six  instruction-tuned LLMs in our experiment, encompassing both commercial APIs and open-source models. From the commercial side, our selection included {PaLM 2} \cite{Anil2023PaLM2T}, Gemini Pro (\texttt{gemini-pro}) \cite{Anil2023GeminiAF}, and ChatGPT (\texttt{gpt-3.5-turbo-1106}) \cite{OpenAIChatGPT}. For open-source models, we employ {LLaMA 2} \cite{touvron2023llama} with different model sizes (\texttt{Llama-2-chat-7b/13b/70b}).

\subsection{Notaions} \label{notation}

For a given question \(q\), the number of options available is denoted by \(k\). Each option within this range, from position \(1\) to \(k\), is characterized by an option symbol \(s_i\) and the corresponding option content \(c_i\), where \(s_i \in S_q\) and \(c_i \in C_q\). Here, \(S_q\) denotes the option symbol set, and \(C_q\) represents the option content set. For instance, consider a question \(q\) that offers four possible answers with the symbol set \(S_q = \{s_1, s_2, s_3, s_4\}\) and \(C_q = \{c_1, c_2, c_3, c_4\}\); in this scenario, the representation of \(q\) can be expressed as \(q = \{(s_1, c_1), (s_2, c_2), (s_3, c_3), (s_4, c_4)\}\).

\subsection{Other Details}
Following HuggingFace Open LLM Leaderboard \cite{open-llm-leaderboard}, we utilize the EleutherAI lm-harness \cite{eval-harness} tool to manage datasets for our experiments. For commercial APIs, we set the temperature to 0 to guarantee reproducibility. For open-source models, we employ Azure AI Studio to deploy various sizes of \texttt{Llama-2-Chat} for parallel processing, optimizing our experimental setup for efficiency and scalability. Additionally, all experiments in this study are conducted in the zero-shot setting, with the prompts being consistent with those used in prior research \cite{Zheng2023JudgingLW,Wang2023LargeLM}. Details of the prompts can be found in Appendix \ref{sec:prompt_format}.

\begin{table*}
\centering
\small
\begin{tabular}{c|cc|cc|cc|cc|cc|cc}
\hline
\multicolumn{1}{c|}{\textbf{Model/}} & \multicolumn{2}{c|}{\textbf{ARC}} & \multicolumn{2}{c|}{\textbf{HellaSwag}} & \multicolumn{2}{c|}{\textbf{MMLU}} & \multicolumn{2}{c|}{\textbf{Winogrande}} & \multicolumn{2}{c|}{\textbf{MathQA}} & \multicolumn{2}{c}{\textbf{OpenBookQA}} \\
\multicolumn{1}{c|}{\textbf{Setting}}& \multicolumn{2}{c|}{\textbf{Acc} / \textbf{Fluct.}} &  \multicolumn{2}{c|}{\textbf{Acc} / \textbf{Fluct.}} &  \multicolumn{2}{c|}{\textbf{Acc} / \textbf{Fluct.}} &  \multicolumn{2}{c|}{\textbf{Acc} / \textbf{Fluct.}} &  \multicolumn{2}{c|}{\textbf{Acc} / \textbf{Fluct.}} & \multicolumn{2}{c}{\textbf{Acc} / \textbf{Fluct.}} \\
\hline
PaLM 2/T & \multicolumn{2}{c|}{82.15 / {\textcolor{blue}{4.98}}} & \multicolumn{2}{c|}{91.06 / \textcolor{blue}{4.82}} & \multicolumn{2}{c|}{64.32 / \textcolor{blue}{15.94}} & \multicolumn{2}{c|}{67.48 / 23.92}& \multicolumn{2}{c|}{30.87 / \textcolor{blue}{36.23}} & \multicolumn{2}{c}{84.7 / \textcolor{blue}{4.2}} \\
PaLM 2/O & \multicolumn{2}{c|}{81.29 / 14.42} & \multicolumn{2}{c|}{90.85 / \textcolor{red}{10.19}} & \multicolumn{2}{c|}{63.70 / 25.59} & \multicolumn{2}{c|}{72.93 / \textcolor{blue}{10.34}} & \multicolumn{2}{c|}{30.18 / \textcolor{red}{67.59}}& \multicolumn{2}{c}{85.40 / 9.00} \\
PaLM 2/B & \multicolumn{2}{c|}{82.32 / \textcolor{red}{14.60}} & \multicolumn{2}{c|}{92.12 / 7.47} & \multicolumn{2}{c|}{63.46 / \textcolor{red}{32.08}} & \multicolumn{2}{c|}{68.07 / \textcolor{red}{34.58}}& \multicolumn{2}{c|}{30.55 / 58.68}& \multicolumn{2}{c}{86.40 / \textcolor{red}{9.24}}\\
\hline
Gemini Pro/T & \multicolumn{2}{c|}{85.15 / \textcolor{blue}{5.67}} & \multicolumn{2}{c|}{79.09 / \textcolor{blue}{15.97}} & \multicolumn{2}{c|}{65.75 / \textcolor{blue}{18.99}} & \multicolumn{2}{c|}{61.29 / \textcolor{blue}{15.07}} & \multicolumn{2}{c|}{26.38 / \textcolor{blue}{34.71}} & \multicolumn{2}{c}{83.10 / \textcolor{blue}{8.20}} \\
Gemini Pro/O & \multicolumn{2}{c|}{84.51 / \textcolor{red}{15.71}} & \multicolumn{2}{c|}{79.04 / 22.55} & \multicolumn{2}{c|}{64.80 / 32.10} & \multicolumn{2}{c|}{60.46 / 45.62} & \multicolumn{2}{c|}{26.31 / \textcolor{red}{66.50}} & \multicolumn{2}{c}{82.0 / \textcolor{red}{19.80}} \\
Gemini Pro/B & \multicolumn{2}{c|}{84.42 / \textcolor{red}{15.71}} & \multicolumn{2}{c|}{78.77 / \textcolor{red}{23.46}} & \multicolumn{2}{c|}{64.38 / \textcolor{red}{36.29}} & \multicolumn{2}{c|}{60.46 / \textcolor{red}{61.56}} & \multicolumn{2}{c|}{26.65 / 71.56} & \multicolumn{2}{c}{83.40 / 19.00} \\
\hline
GPT 3.5/T & \multicolumn{2}{c|}{75.24 / \textcolor{blue}{15.87}} & \multicolumn{2}{c|}{78.74 / \textcolor{blue}{14.54}} & \multicolumn{2}{c|}{58.29 / \textcolor{blue}{24.20}} & \multicolumn{2}{c|}{54.46 / \textcolor{blue}{22.08}} & \multicolumn{2}{c|}{14.07 / \textcolor{blue}{28.19}} & \multicolumn{2}{c}{71.90 / \textcolor{blue}{15.20}} \\
GPT 3.5/O & \multicolumn{2}{c|}{75.79 / \textcolor{red}{19.62}} & \multicolumn{2}{c|}{78.76 / 18.73} & \multicolumn{2}{c|}{58.36 / \textcolor{red}{31.01}} & \multicolumn{2}{c|}{54.97 / 29.83} & \multicolumn{2}{c|}{14.20 / 30.94} & \multicolumn{2}{c}{70.60 / \textcolor{red}{26.40}} \\
GPT 3.5/B & \multicolumn{2}{c|}{77.98 / 17.94} & \multicolumn{2}{c|}{78.69 / \textcolor{red}{19.57}} & \multicolumn{2}{c|}{59.36 / 28.76} & \multicolumn{2}{c|}{54.50 / \textcolor{red}{40.51}} & \multicolumn{2}{c|}{12.83 / \textcolor{red}{62.15}} & \multicolumn{2}{c}{73.70 / 22.29}\\
\hline
LLaMA2-7B/T & \multicolumn{2}{c|}{38.07 / \textcolor{blue}{53.20}}& \multicolumn{2}{c|}{39.21 / \textcolor{blue}{57.2}} & \multicolumn{2}{c|}{32.22 / \textcolor{blue}{51.51}}& \multicolumn{2}{c|}{46.65 / \textcolor{blue}{4.27}}& \multicolumn{2}{c|}{15.31 / 61.35}& \multicolumn{2}{c}{29.60 / \textcolor{blue}{62.45}}\\
LLaMA2-7B/O & \multicolumn{2}{c|}{37.38 / \textcolor{red}{71.43}}& \multicolumn{2}{c|}{39.30 / \textcolor{red}{63.03}}& \multicolumn{2}{c|}{30.38 / 66.41}& \multicolumn{2}{c|}{47.00 / 96.57}& \multicolumn{2}{c|}{16.18 / \textcolor{blue}{56.80}}& \multicolumn{2}{c}{32.90 / \textcolor{red}{82.73}}\\
LLaMA2-7B/B & \multicolumn{2}{c|}{39.31 / 68.13}& \multicolumn{2}{c|}{41.17 / 60.54}& \multicolumn{2}{c|}{31.42 / \textcolor{red}{74.53}}& \multicolumn{2}{c|}{46.72 / \textcolor{red}{100.00}}& \multicolumn{2}{c|}{16.89 / \textcolor{red}{70.98}}& \multicolumn{2}{c}{33.70 / 75.40}\\
\hline
LLaMA2-13B/T & \multicolumn{2}{c|}{45.62 / 38.64}& \multicolumn{2}{c|}{38.02 / \textcolor{blue}{36.62}}& \multicolumn{2}{c|}{36.96 / \textcolor{blue}{38.90}}& \multicolumn{2}{c|}{44.00 / 88.67}& \multicolumn{2}{c|}{18.91 / 48.01}& \multicolumn{2}{c}{37.70 / 48.04}\\
LLaMA2-13B/O & \multicolumn{2}{c|}{45.97 / \textcolor{blue}{36.29}}& \multicolumn{2}{c|}{38.11 / \textcolor{red}{54.46}}& \multicolumn{2}{c|}{36.67 / 39.18}& \multicolumn{2}{c|}{43.80 / \textcolor{blue}{3.84}}& \multicolumn{2}{c|}{18.53 / \textcolor{blue}{47.08}}& \multicolumn{2}{c}{39.40 / \textcolor{blue}{37.14}}\\
LLaMA2-13B/B & \multicolumn{2}{c|}{46.18 / \textcolor{red}{45.55}} & \multicolumn{2}{c|}{37.32 / 52.44}& \multicolumn{2}{c|}{36.78 / \textcolor{red}{54.73}}& \multicolumn{2}{c|}{45.42 / \textcolor{red}{99.56}}& \multicolumn{2}{c|}{19.77 / \textcolor{red}{76.17}}& \multicolumn{2}{c}{41.90 / 48.43}\\
\hline
LLaMA2-70B/T & \multicolumn{2}{c|}{60.17 / \textcolor{red}{37.40}}& \multicolumn{2}{c|}{58.66 / \textcolor{red}{52.94}}& \multicolumn{2}{c|}{44.95 / \textcolor{red}{55.06}}& \multicolumn{2}{c|}{47.71 / 96.30}& \multicolumn{2}{c|}{23.13 / 73.70}& \multicolumn{2}{c}{58.00 / \textcolor{red}{50.30}}\\
LLaMA2-70B/O & \multicolumn{2}{c|}{60.17 / \textcolor{blue}{35.88}}& \multicolumn{2}{c|}{58.85 / 50.80}& \multicolumn{2}{c|}{46.29 / 49.62}& \multicolumn{2}{c|}{48.62 / \textcolor{blue}{20.08}}& \multicolumn{2}{c|}{23.25 / \textcolor{red}{82.04}}& \multicolumn{2}{c}{55.10 / 44.78}\\
LLaMA2-70B/B & \multicolumn{2}{c|}{61.37 / {35.16}}& \multicolumn{2}{c|}{64.42 / \textcolor{blue}{27.77}}& \multicolumn{2}{c|}{47.32 / \textcolor{blue}{42.57}}& \multicolumn{2}{c|}{47.59 / \textcolor{red}{100.00}}& \multicolumn{2}{c|}{24.54 / \textcolor{blue}{37.82}}& \multicolumn{2}{c}{60.40 / \textcolor{blue}{38.60}}\\
\hline
\end{tabular}
\caption{\label{sensitivity-results}
Results of model sensitivity experiment across models and tasks, with sensitivity settings denoted as \textbf{T} (Token), \textbf{O} (Order), and \textbf{B} (Both). \textbf{Acc} (\%) represents the mean of \(r_{forward}\) and \(r_{backward}\) accuracies for each setting. For each model, the minimum fluctuation rate is highlighted in \textcolor{blue}{blue}, signifying lower sensitivity and the maximum rate is marked in \textcolor{red}{red}, indicating higher sensitivity.}
\end{table*}

\section{Investigation on LLM Sensitivity}
\label{investigate_llm_sensitivitiy}

While prior research has touched upon biases in LLMs concerning MCQs, with notable findings on position bias and token bias, our work stands out by delving deeper into unexplored territories of the combined impact of option order and token usage within MCQs. We uncover novel insights into the decision-making processes of LLMs that have yet to be extensively explored in the existing literature. 

\subsection{Setups}
We adhere to the notations established in Section \ref{notation}, allowing for a more coherent and precise description of the experimental setups.

\noindent \textbf{Token Sensitivity.}
To assess the impact of token sensitivity, we employ the default option symbol set \(S_q = \{A, B, \ldots, S_{qk}\}\) for each question \(q\), where \(k\) represents the number of option contents for question \(q\) and and \(S_{qk}\) represents the \(k\)-th letter of the alphabet from A to Z. For each question, we conduct experiments with two distinct requests to the LLM. The first request is defined as follows:
\begin{equation}
r_{forward} = \{(s_i, c_i) \mid i = 1, 2, \ldots, k\}
\end{equation}

\noindent Here,  \(s_{i}\) refers to the \(i\)-th option symbol in \(S_q\), and \(c_i\) represents the \(i\)-th option content of \(C_q\), indicating the \(i\)-th answer candidate for the question.

Conversely, the second request, \(r_{backward}\), introduces a reversed  arrangement of the option symbols, as detailed below:
\begin{equation}
r_{backward} = \{(s_{k-i+1}, c_{i}) \mid i = 1, 2, \ldots, k\}
\end{equation}
Subsequently, the results of \(r_{forward}\) and \(r_{backward}\) are analyzed. 

\noindent \textbf{Order Sensitivity.}
To determine the influence of order sensitivity, we adopt a strategy of coupling each option symbol with its corresponding option content, thereby aiming to nullify the effects of token sensitivity. Consistent with the settings described previously, the option symbol set \(S_q\) is \(\{A, B, \ldots, S_{qk}\}\). \(r_{forward}\) and \(r_{backward}\) are:
\begin{equation}
r_{forward} = \{(s_i, c_i) \mid i = 1, 2, \ldots, k\}
\end{equation}
\begin{equation}
r_{backward} = \{(s_i, c_i) \mid i = k, k-1, \ldots, 1\}
\end{equation}

\noindent \textbf{Both Sensitivity.}
In practical scenarios, a common remediation strategy involves rearranging the order of option content. This maneuver inherently addresses both order and token sensitivities. It is anticipated that if the biases induced by these sensitivities align, their cumulative effect on sensitivity will be magnified. Conversely, if they are in opposition, their effects will likely be mitigated. Following the previously described setting, where the symbol set \(S_q\) is \(\{A, B, \ldots, S_{qk}\}\), we define \(r_{forward}\) and \(r_{backward}\) as follows:

\begin{equation}
r_{forward} = \{(s_i, c_i) \mid i = 1, 2, \ldots, k\}
\end{equation}
\begin{equation}
r_{backward} = \{(s_i, c_{k-i+1}) \mid i = 1, 2, \ldots, k\}
\end{equation}

\subsection{Measurement of Sensitivity} \label{flu-rate}
To assess the model's sensitivity, we introduce the \textit{Fluctuation Rate} (FR), a metric designed to quantify the variability in responses between \(r_{forward}\) and \(r_{backward}\).  The equation for \(FR\) is given by:
\begin{equation}
FR =\frac{\sum_{i=1}^{N} (r_{forward}(i) \neq r_{backward}(i))}{N}
\end{equation}
where \(\sum_{i=1}^{N} (r_{forward}(i) \neq r_{backward}(i))\) denotes the number of instances where the outcomes of \(r_{forward}\) and \(r_{backward}\) are not identical, and \(N\) represents the sample size of that task. Thus, \(FR\) reflects the fraction of all questions where the two requests yield divergent results.

\subsection{Overall Observation} 
\label{sensitivity-result-descibtion}
Table \ref{sensitivity-results} comprehensively summarizes our sensitivity experiments across various LLMs. 
We also provide detailed breakdowns of the MMLU's performance across its 57 subtasks in Appendix \ref{detailed-sensitivity-result}. 
In powerful LLMs, PaLM 2, Gemini Pro, and GPT-3.5, we observe a notable trend: they are more sensitive to option order than to symbols/tokens in 17 out of 18 cases.
An exception to this trend is observed with the Winogrande dataset, where PaLM 2 shows increased sensitivity to token variations. 
In \textit{both sensitivity} setting, which examines the joint effects of token and order sensitivities, we find that in 11 out of 18 cases, the combined influence is the most pronounced. This indicates that in more than half of the cases, the directional impacts of token and order sensitivities tend to align.

Conversely, the open-source LLM, LLaMA 2 (\texttt{Llama-2-chat}), across its varying sizes, does not exhibit a consistent sensitivity trend towards token and order. For instance, while the 7B model appears more sensitive to order, the 13B and 70B models do not follow this pattern. Although Table \ref{sensitivity-results} indicates that the 13B and 70B models are more sensitive to token differences in 9 out of 12 instances, the discrepancy in the fluctuation rate between token and order sensitivity is marginal.

\begin{figure*}[htbp]
\centering
\includegraphics[width=\linewidth]{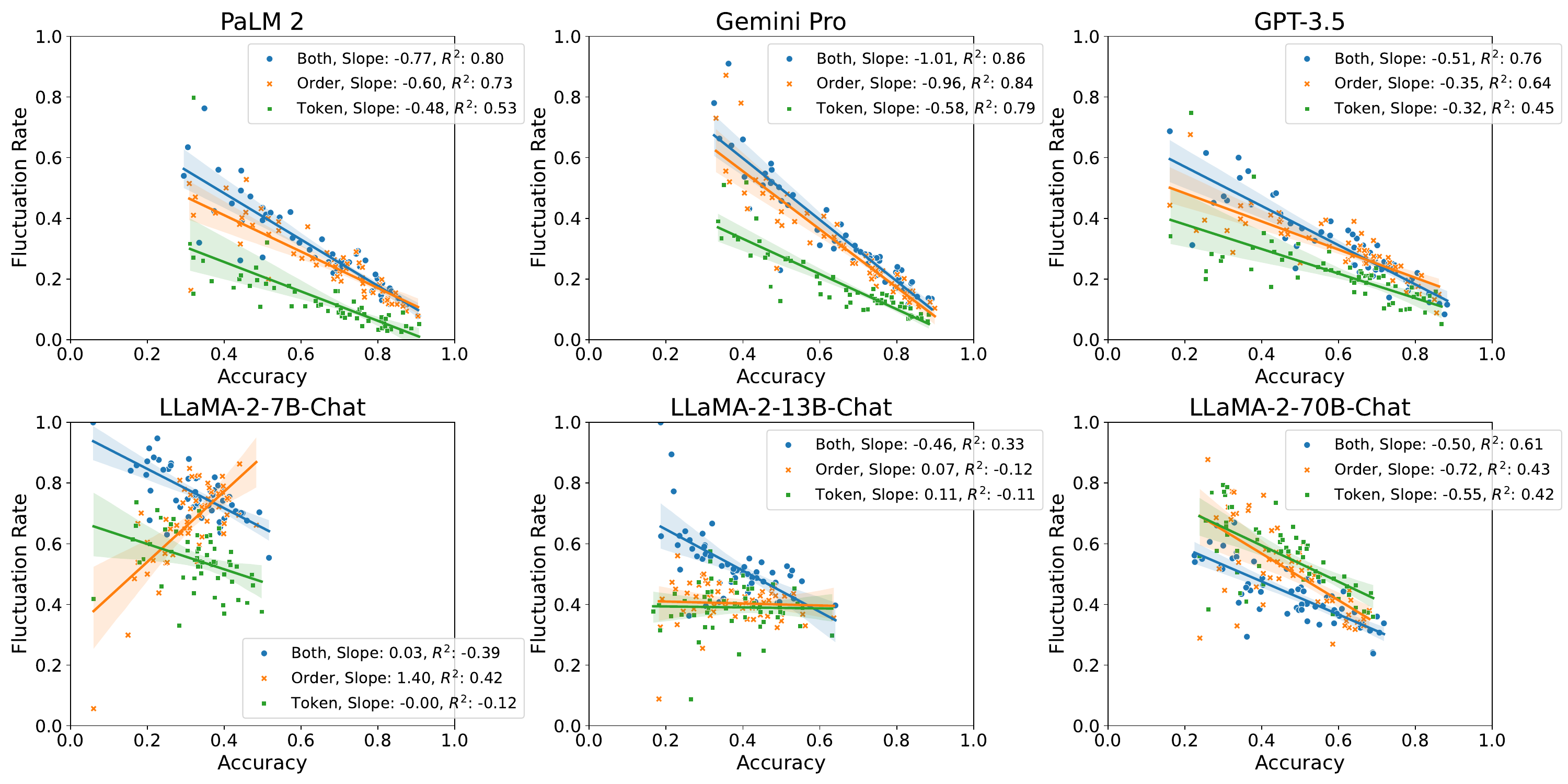}
\caption{
Correlation between model accuracy and fluctuation rates under different sensitivity settings: \textit{Token}, \textit{Order}, and \textit{Both}. Including linear regression lines for each setting, alongside slope and $R^2$ values, to clearly show the relation between model performance and fluctuation rates.
}
\label{fig:model_sensitivity}
\end{figure*}

\subsection{Relation between Difficulty and Sensitivity}
Table \ref{sensitivity-results} reveals an interesting pattern: tasks with higher accuracy levels, such as ARC Challenge, HellaSwag, and OpenBookQA, tend to exhibit lower fluctuation rates. This observation prompts us to question whether there is a relationship between the difficulty of a task and the sensitivity of a model to it. To further investigate this hypothesis, we analyze the sensitivity across 57 MMLU subtasks. For detailed results per model, we refer to Appendix \ref{detailed-sensitivity-result}, as mentioned in Section \ref{sensitivity-result-descibtion}, due to space constraints.

Figure \ref{fig:model_sensitivity} illustrates the correlation between task accuracy and fluctuation rates across six models, encompassing three advanced commercial LLMs and three open-source models of varying sizes. This comparison offers a unique opportunity to assess the impact of scaling model parameters. For comprehensive insight, we integrate the previously discussed settings—\textit{token sensitivity}, \textit{order sensitivity}, and \textit{both sensitivity}—into a single diagram per model, facilitating a clear understanding of how task difficulty correlates with model sensitivity.

Results from PaLM 2, Gemini Pro, GPT-3.5, and LLaMA 2 70B appear to support our hypothesis: more challenging tasks, characterized by lower accuracy, tend to exhibit greater sensitivity, as indicated by higher fluctuation rates. This aligns with our intuition that models are more confident and thus less sensitive to fluctuations in easier questions. 
A notable observation pertains to the smaller LLaMA 2 models, specifically the 7B and 13B versions. Tasks that are more straightforward for other powerful LLMs pose significant challenges to these models, leading to lower accuracy and a muted trend in sensitivity as tasks vary in difficulty. However, a closer analysis of the 7B, 13B, and 70B models reveals a gradual manifestation of the expected trend. The shift from the 7B to the 13B model, for instance, corresponds with our expectation in the \textit{both sensitivity} setting. With further increases in model size, the 70B model exhibits the predicted correlation between task difficulty and model sensitivity across all examined settings.

\begin{table}[h!]
\centering
\small
\begin{tabular}{l|c|c|c|c}
\hline
 & A (\%) & B (\%) & C (\%) & D (\%) \\
\hline
Ground truth & 22.58 & 26.52 & \textbf{26.52} & 24.38 \\
\hline
PaLM 2 & 18.30 & 26.29 & \textbf{28.69} & 26.72 \\
Gemini Pro & 18.03 & 27.81 & \textbf{29.10} & 25.06 \\
GPT 3.5 & 18.46 & 29.48 & \textbf{30.18} & 21.87 \\
LLaMA2-7B & \textbf{57.39} & 19.98 & 22.62 & 0.00 \\
LLaMA2-13B & 1.41 & 42.03 & \textbf{43.44} & 13.13 \\
LLaMA2-70B & 7.23 & 31.78 & \textbf{41.67} & 19.32 \\
\hline
\end{tabular}
\caption{\label{proportion_proportion} Option proportion statistics and ground truth label proportions for the ARC dataset. The most frequent option in each row is highlighted in \textbf{bold}.}
\end{table}

\subsection{LLMs' Option Tendency}
To understand LLMs' behavior, we calculated the option proportion statistics to analyze their tendencies. Specifically, we calculated the answer distribution of \(r_{forward}\) and detailed the information for each option alongside the ground truth label proportion. Table \ref{proportion_proportion} shows the results for the ARC dataset, highlighting the similarities and differences in selection biases among various LLMs. According to the results, most models, except for LLaMA2-7B, exhibit a notable bias towards option C compared to the ground truth proportion.
Due to space limitations, statistics for the other five datasets, including HellaSwag, MMLU, Winogrande, MathQA, and OpenBookQA, are included in Appendix \ref{llm-option-stats}. Generally, most models, except for LLaMA2-7B, show a bias towards options B or C.

\section{Methodology}
To mitigate the impact of sensitivity to tokens and/or order and improve model stability, we propose three methods tailored to different contexts of LLMs. We categorize these contexts into two scenarios: Gray-Box and Black-Box. In a Black-Box scenario, the LLM provides only the generated text upon request, without additional information. Conversely, a Gray-Box scenario allows access to more detailed output, such as token probability information. In our experiments, GPT-3.5 falls into the Gray-Box category, as the OpenAI API enables retrieval of the top 5 token log probabilities, whereas other models are Black-Box ones. Furthermore, all experiments adhere to the sensitivity settings mentioned in Section \ref{investigate_llm_sensitivitiy}.

\subsection{Gray-Box Probability Weighting}
\label{gray-box-weighted-section}
For each question \(q\), the requests \(r_{forward}\) and \(r_{backward}\) are:
\begin{equation}
r_{forward} = \{(s^f_i, c^f_i) \mid i = 1, 2, \ldots, k\}
\end{equation}
\begin{equation}
r_{backward} = \{(s^b_j, c^b_j) \mid j = 1, 2, \ldots, k\}
\end{equation}

Let function \(p(\cdot)\) represents the probability of a token generated by the LLM. For instance, \(p(s^{f}_{3})\) represents the probability that the model selects the third option symbol of \(r_{forward}\).
We calculate the weighted probability for each option content \(c^f_i\) in the first query set \(r_{forward}\). The weighted probability of a specific option content \(c^f_i\) is derived by integrating the probabilities of its corresponding symbol \(s^f_i\) in \(r_{forward}\) with the symbol \(s^b_j\) in the second query set \(r_{backward}\). The formulation of this computation is as follows:
\begin{equation}
P^{weighted}_{c^f_i} = p(s^f_i) \times p(s^b_j) \quad \text{where }  c^f_i = c^b_j
\end{equation}

The final choice is \(c^f_{i^*}\), the option content with the highest weighted probability, determined by:
\begin{equation}
i^* = \underset{i}{\text{argmax}} \, P^{weighted}_{c^f_i},
\end{equation}
where \({f^*}\) maximizes \(P^{weighted}_{c^f_i}\).

\begin{table*}[t]
\centering
\small
\begin{tabular}{c|c|c|c|c|c|c|c}
\hline
  \textbf{Method}&\textbf{Setting} &  \textbf{ARC}&\textbf{HellaSwag}&\textbf{MMLU}&  \textbf{Winogrande}& \textbf{MathQA}&\textbf{OpenBookQA}\\
\hline
 &Token& \textbf{77.60} \textcolor{blue}{\scriptsize(+2.36)}&  \textbf{79.94} \textcolor{blue}{\scriptsize(+1.20)}& \textbf{60.52} \textcolor{blue}{\scriptsize(+2.23)}&56.67 \textcolor{blue}{\scriptsize(+2.21)}& 17.09 \textcolor{blue}{\scriptsize(+3.02)}& \textbf{75.00} \textcolor{blue}{\scriptsize(+3.10)}\\
 Weighting&Order& \textbf{78.80} \textcolor{blue}{\scriptsize(+3.00)}&  \textbf{80.58} \textcolor{blue}{\scriptsize(+1.82)}& \textbf{60.93} \textcolor{blue}{\scriptsize(+2.57)}&59.04 \textcolor{blue}{\scriptsize(+4.06)}& 17.42 \textcolor{blue}{\scriptsize(+3.22)}& \textbf{75.00} \textcolor{blue}{\scriptsize(+4.40)}\\
 &Both& \textbf{80.26} \textcolor{blue}{\scriptsize(+2.27)}&  \textbf{81.27} \textcolor{blue}{\scriptsize(+2.58)}& \textbf{61.83} \textcolor{blue}{\scriptsize(+2.48)}& \textbf{59.12} \textcolor{blue}{\scriptsize(+4.62)}& 16.08 \textcolor{blue}{\scriptsize(+3.25)}& \textbf{77.80} \textcolor{blue}{\scriptsize(+4.10)}\\
\hline
&Token& 75.62 \textcolor{blue}{\scriptsize(+0.39)}&  79.35 \textcolor{blue}{\scriptsize(+0.61)}& 59.49 \textcolor{blue}{\scriptsize(+1.20)}& \textbf{59.31} \textcolor{blue}{\scriptsize(+4.85)}& \textbf{23.79} \textcolor{blue}{\scriptsize(+9.72)}& 72.40 \textcolor{blue}{\scriptsize(+0.50)}\\
Calibration&Order& 76.22 \textcolor{blue}{\scriptsize(+0.43)}&  79.32 \textcolor{blue}{\scriptsize(+0.56)}& 59.41 \textcolor{blue}{\scriptsize(+1.05)}& \textbf{59.71} \textcolor{blue}{\scriptsize(+4.74)}& \textbf{22.60} \textcolor{blue}{\scriptsize(+8.39)}& 70.60 {\scriptsize(+0.00)}\\
&Both& 78.07 \textcolor{blue}{\scriptsize(+0.09)}&  79.30 \textcolor{blue}{\scriptsize(+0.61)}& 60.62 \textcolor{blue}{\scriptsize(+1.27)}&58.84 \textcolor{blue}{\scriptsize(+4.34)}& \textbf{22.41} \textcolor{blue}{\scriptsize(+9.58)}& 74.30 \textcolor{blue}{\scriptsize(+0.60)}\\
\hline
\end{tabular}
\caption{\label{graybox-results}
Results of gray-box methods of GPT-3.5 model. Accuracy is presented in percentage format, with the highest results in each setting \textbf{bolded}. Differences from the original results are shown in parentheses, with positive improvements highlighted in \textcolor{blue}{blue}, indicating enhanced performance following our method.
}
\end{table*}

\subsection{Gray-Box Probability Calibration}
Due to the sensitivities of LLMs to both the order and tokens in MCQs, their outputs frequently show biases, leading to preferences for specific options. To address this issue and promote a fairer and more accurate answer selection process, we calibrate the output probabilities. This calibration aims to enhance the precision of which the model selects answers.

Let the output distributions for each option symbol in \(r_{forward}\) and \(r_{backward}\) are denoted by \(D_{forward}\) and \(D_{backward}\), respectively, and are formulated as follows:
\begin{equation}
D_{forward} = \{p_d(s^f_i) \mid i = 1, 2, \ldots, k\}
\end{equation}
\begin{equation}
D_{backward} = \{p_d(s^b_j) \mid j = 1, 2, \ldots, k\}
\end{equation}
where \(p_d(s_i)\) represents the probability distribution of option symbol \(s_i\)\ , defined by:
\begin{equation}
p_d(s_i) = \frac{\text{count}(s_i)}{N}
\end{equation}

Here, \(N\) denotes the total sample count, and \(\text{count}(s_i)\) indicates the number of samples for which the model selects \(s_i\) as the answer. Thus, \(p_d(s_i)\) reflects the percentage of selections for \(s_i\). For real-world applicability, we calculate these distributions using the validation set of each task.

To calculate the calibrated probabilities, we use the following formulations:
\begin{equation}
P^{calibrated}_{forward} = \left\{\frac{p(s^f_i)}{p_d(s^f_i)} \mid i = 1, 2, \ldots, k\right\}
\end{equation}
\begin{equation}
P^{calibrated}_{backward} = \left\{\frac{p(s^b_j)}{p_d(s^b_j)} \mid j = 1, 2, \ldots, k\right\}
\end{equation}

Here, \(P^{calibrated}_{forward}\) and \(P^{calibrated}_{backward}\) represent the sets of calibrated probabilities for each option symbol in \(r_{forward}\) and \(r_{backward}\), respectively. The calibration process of \(P^{calibrated}_{forward}\) involves dividing the original probability of selecting each symbol \(p(s^f_i)\) by its corresponding output distribution probability \(p_d(s^f_i)\), for \(i = 1, 2, \ldots, k\). This approach ensures that each option's probability is adjusted in light of its observed selection frequency, aiming to align the model's output more closely with an unbiased selection criterion.

Considering the three distinct sensitivity settings, we identify three specific distribution sets: (\(D^{token}_{forward}\), \(D^{token}_{backward}\)), (\(D^{order}_{forward}\), \(D^{order}_{backward}\)), and (\(D^{both}_{forward}\), \(D^{both}_{backward}\)). These distributions underpin our calibration strategy, allowing us to adjust the model's outputs to reduce bias and enhance answer accuracy across different sensitivity contexts.

\begin{table*}[t]
\centering
\small
\begin{tabular}{c|c|c|c|c|c|c|c}
\hline
\textbf{Model}& \textbf{Setting} &  \textbf{ARC}&\textbf{HellaSwag}&\textbf{MMLU}&  \textbf{Winogrande}& \textbf{MathQA}&\textbf{OpenBookQA}\\
\hline
&Token& 82.15 {\scriptsize(+0.00)}&  91.46 \textcolor{blue}{\scriptsize(+0.39)}& 64.52 \textcolor{blue}{\scriptsize(+0.20)}&66.54 \textcolor{red}{\scriptsize(-0.95)}& 31.52 \textcolor{blue}{\scriptsize(+0.65)}& 85.40 \textcolor{blue}{\scriptsize(+0.70)}\\
PaLM 2&Order& 81.63 \textcolor{blue}{\scriptsize(+0.34)}&  91.55 \textcolor{blue}{\scriptsize(+0.69)}& 64.09 \textcolor{blue}{\scriptsize(+0.39)}&71.43 \textcolor{red}{\scriptsize(-1.50)}& 31.69 \textcolor{blue}{\scriptsize(+1.51)}& 85.60 \textcolor{blue}{\scriptsize(+0.20)}\\
&Both& 83.00 \textcolor{blue}{\scriptsize(+0.69)}&  92.24 {\scriptsize(+0.12)}& 64.05 \textcolor{blue}{\scriptsize(+0.59)}&62.27 \textcolor{red}{\scriptsize(-5.80)}& 31.62 \textcolor{blue}{\scriptsize(+1.07)}& 86.20 \textcolor{red}{\scriptsize(-0.20)}\\
\hline
&Token& 85.67 \textcolor{blue}{\scriptsize(+0.52)}&  78.77 \textcolor{red}{\scriptsize(-0.32)}& 65.91 \textcolor{blue}{\scriptsize(+0.16)}&62.43 \textcolor{blue}{\scriptsize(+1.14)}& 27.27 \textcolor{blue}{\scriptsize(+0.89)}& 83.00 \textcolor{red}{\scriptsize(-0.10)}\\
Gemini Pro&Order& 85.84 \textcolor{blue}{\scriptsize(+1.33)}&  79.70 \textcolor{blue}{\scriptsize(+0.66)}& 65.75 \textcolor{blue}{\scriptsize(+0.95)}&59.43 \textcolor{red}{\scriptsize(-1.03)}& 26.37 \textcolor{blue}{\scriptsize(+0.05)}& 83.20 \textcolor{blue}{\scriptsize(+0.80)}\\
&Both& 85.67 \textcolor{blue}{\scriptsize(+1.24)}&  80.31 \textcolor{blue}{\scriptsize(+1.54)}& 65.88 \textcolor{blue}{\scriptsize(+1.50)}&59.27 \textcolor{red}{\scriptsize(-1.18)}& 26.67 \textcolor{blue}{\scriptsize(+0.02)}& 84.40 \textcolor{blue}{\scriptsize(+1.00)}\\
\hline
&Token& 76.74 \textcolor{blue}{\scriptsize(+1.50)}&  80.90 \textcolor{blue}{\scriptsize(+2.16)}& 59.69 \textcolor{blue}{\scriptsize(+1.40)}&53.99 \textcolor{red}{\scriptsize(-0.47)}& 12.23 \textcolor{red}{\scriptsize(-1.84)}& 72.60 \textcolor{blue}{\scriptsize(+0.70)}\\
GPT-3.5&Order& 76.39 \textcolor{blue}{\scriptsize(+0.60)}&  81.29 \textcolor{blue}{\scriptsize(+2.53)}& 59.44 \textcolor{blue}{\scriptsize(+1.08)}&51.62 \textcolor{red}{\scriptsize(-3.35)}& 12.09 \textcolor{red}{\scriptsize(-2.11)}& 72.00 \textcolor{blue}{\scriptsize(+1.40)}\\
&Both& 78.03 \textcolor{blue}{\scriptsize(+0.04)}&  80.81 \textcolor{blue}{\scriptsize(+2.12)}& 60.47 \textcolor{blue}{\scriptsize(+1.12)}&49.25 \textcolor{red}{\scriptsize(-5.25)}& 11.56 \textcolor{red}{\scriptsize(-1.27)}& 73.80 \textcolor{blue}{\scriptsize(+0.10)}\\
\hline
&Token& 41.29 \textcolor{blue}{\scriptsize(+3.22)}&  37.46 \textcolor{red}{\scriptsize(-1.75)}& 32.27 \textcolor{blue}{\scriptsize(+0.05)}&45.94 \textcolor{red}{\scriptsize(-0.71)}& 15.61 \textcolor{blue}{\scriptsize(+0.30)}& 32.60 \textcolor{blue}{\scriptsize(+3.00)}\\
LLaMA2-7B&Order& 41.29 \textcolor{blue}{\scriptsize(+3.91)}&  47.24 \textcolor{blue}{\scriptsize(+7.94)}& 32.35 \textcolor{blue}{\scriptsize(+1.97)}&46.72 \textcolor{red}{\scriptsize(-0.28)}& 15.61 \textcolor{red}{\scriptsize(-0.57)}& 32.60 \textcolor{red}{\scriptsize(-0.30)}\\
&Both& 41.29 \textcolor{blue}{\scriptsize(+1.97)}&  45.02 \textcolor{blue}{\scriptsize(+3.85)}& 32.32 \textcolor{blue}{\scriptsize(+0.91)}&46.33 \textcolor{red}{\scriptsize(-0.39)}& 15.61 \textcolor{red}{\scriptsize(-1.27)}& 32.60 \textcolor{red}{\scriptsize(-1.10)}\\
\hline
&Token& 45.24 \textcolor{red}{\scriptsize(-0.39)}&  41.63 \textcolor{blue}{\scriptsize(+3.61)}& 39.19 \textcolor{blue}{\scriptsize(+2.23)}&42.30 \textcolor{red}{\scriptsize(-1.70)}& 19.13 \textcolor{blue}{\scriptsize(+0.22)}& 35.20 \textcolor{red}{\scriptsize(-2.50)}\\
LLaMA2-13B&Order& 47.47 \textcolor{blue}{\scriptsize(+1.50)}&  39.41 \textcolor{blue}{\scriptsize(+1.30)}& 37.74 \textcolor{blue}{\scriptsize(+1.07)}&41.75 \textcolor{red}{\scriptsize(-2.05)}& 18.76 \textcolor{blue}{\scriptsize(+0.23)}& 43.20 \textcolor{blue}{\scriptsize(+3.80)}\\
&Both& 48.76 \textcolor{blue}{\scriptsize(+2.58)}&  39.17 \textcolor{blue}{\scriptsize(+1.84)}& 39.40 \textcolor{blue}{\scriptsize(+2.62)}&44.20 \textcolor{red}{\scriptsize(-1.22)}& 19.66 \textcolor{red}{\scriptsize(-0.10)}& 43.20 \textcolor{blue}{\scriptsize(+1.30)}\\
\hline
&Token& 61.55 \textcolor{blue}{\scriptsize(+1.37)}&  62.21 \textcolor{blue}{\scriptsize(+3.55)}& 47.29 \textcolor{blue}{\scriptsize(+2.34)}&47.36 \textcolor{red}{\scriptsize(-0.36)}& 24.36 \textcolor{blue}{\scriptsize(+1.22)}& 57.80 \textcolor{red}{\scriptsize(-0.20)}\\
LLaMA2-70B&Order& 61.03 \textcolor{blue}{\scriptsize(+0.86)}&  58.84 \textcolor{red}{\scriptsize(-0.01)}& 46.12 \textcolor{red}{\scriptsize(-0.17)}&47.36 \textcolor{red}{\scriptsize(-1.26)}& 23.35 \textcolor{blue}{\scriptsize(+0.10)}& 57.00 \textcolor{blue}{\scriptsize(+1.90)}\\
&Both& 64.72 \textcolor{blue}{\scriptsize(+3.35)}&  66.52 \textcolor{blue}{\scriptsize(+2.10)}& 49.03 \textcolor{blue}{\scriptsize(+1.71)}&47.36 \textcolor{red}{\scriptsize(-0.24)}& 23.95 \textcolor{red}{\scriptsize(-0.59)}& 60.80 \textcolor{blue}{\scriptsize(+0.40)}\\
\hline
\end{tabular}
\caption{\label{two-hop-results}
Results of the black-box method. Accuracy is presented in percentage format. Differences from the baseline results are indicated in parentheses: improvements are highlighted in \textcolor{blue}{blue}, signifying enhanced performance due to our method, while declines are marked in \textcolor{red}{red}, indicating a decrease in performance in those scenarios.
}
\end{table*}

\subsection{Black-Box Two-Hop Strategy}
In practical applications, we often encounter black-box scenarios while using commercial LLM APIs. To mitigate the impact of model sensitivity in these situations, we propose a black-box two-hop strategy that leverages the model's output distributions \(D_{forward}\). Given the constraints of black-box scenarios, where recalculating the token probability \(p(s_i)\) is impossible, we adopt an alternative strategy. Our approach intentionally avoids selecting the most biased option symbols in the first request \(r_{forward}\), opting for responses from \(r_{backward}\) instead.
Firstly, we identify the most probable option symbol \(s^f_{i^*}\) based on the distribution \(D_{forward}\), using the equation:
\begin{equation}
i^* = \underset{i}{\text{argmax}} \, p_d(s^f_i),
\end{equation}
where \(p_d(s^f_i)\) denotes the distribution probability of selecting symbol \(s^f_i\) from \(r_{forward}\).
Subsequently, the two-hop strategy is implemented as follows:
\begin{equation}
\text{Final Selection} = 
\begin{cases}
c^f_{j_f^*} & \text{if } i^* \neq j_f^*, \\
c^b_{j_b^*} & \text{if } i^* = j_f^*,
\end{cases}
\end{equation}
where we select \(c^f_{j_f^*}\) from \(r_{forward}\) if it does not exhibit bias most. Otherwise, we opt for \(c^b_{j_b^*}\) as our final answer. Here, \(j_f^*\) and \(j_b^*\) are determined by:
\begin{equation}
j_f^* = \underset{j}{\text{argmax}} \, p(s^f_j),
\end{equation}
\begin{equation}
j_b^* = \underset{j}{\text{argmax}} \, p(s^b_j),
\end{equation}
where \(j_f^*\), \(j_b^*\) indicate the indices of the symbols with the highest probabilities in \(r_{forward}\), \(r_{backward}\), respectively.

This method aims to utilize responses that potentially minimize bias by considering the model's preference patterns indicated by \(D_{forward}\), thereby ensuring the accuracy of selections.

\section{Experiment Results}
\subsection{Gray-Box Results}
Tables \ref{graybox-results} and \ref{two-hop-results} show the results of our methods. In the gray-box scenario, only GPT-3.5 is included since other models do not provide the information of token probability information. Conversely, in the black-box scenario, all models are considered in our experiment. In the subsequent analysis, we compare the performance improvements achieved by our methods against the baseline established in Section \ref{investigate_llm_sensitivitiy}, aiming to underscore the enhancements or limitations observed across tasks and models.

As Table \ref{graybox-results} illustrates, gray-box methods, including both probability weighting and calibration approaches, significantly improve performance across six distinct tasks under three sensitivity settings. Notably, the probability weighting method demonstrates considerable enhancements in all scenarios, surpassing the baseline. It benefits not only more challenging tasks such as MathQA, Winogrande, and MMLU but also shows improvements in easier tasks.
Interestingly, the probability calibration method outperforms the weighting method in two specific tasks out of the six: Winogrande and MathQA. These tasks are unique in their format; Winogrande presents only two options, whereas MathQA offers five options per question. We speculate that the number of options may influence the LLM's preference distribution, thereby affecting the performance of different methods.

\begin{figure*}[t]
\definecolor{darkgreen}{rgb}{0.0, 0.5, 0.0}
\centering
\includegraphics[width=\linewidth]{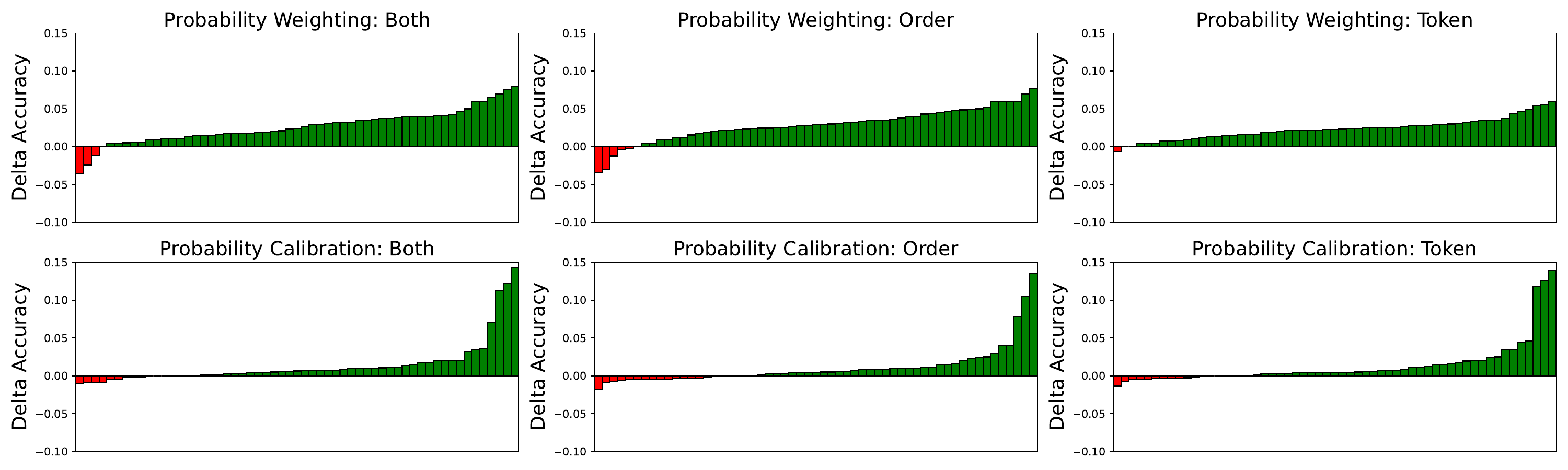}
\caption{
Accuracy Difference Distribution Across 57 MMLU Subtasks For The GPT-3.5 Model in the Gray-Box Scenario: Subtasks are sorted by the difference in accuracy from low to high, indicating that subtasks towards the right benefit more from our methodology. Improvements are marked in \textcolor{darkgreen}{green}, whereas declines in performance are highlighted in \textcolor{red}{red}. The top three diagrams present outcomes from the probability weighting method across three sensitivity settings, while the bottom three diagrams illustrate the effects of the probability calibration method.
}
\label{fig:gray_box_improvement}
\end{figure*}

\subsection{Black-Box Results}
Table \ref{two-hop-results} displays the results of the black-box method applied to various models, sensitivity settings, and tasks. Notably, the stronger models, PaLM 2 and Gemini Pro, show significant benefits from the two-hop strategy. They improved in five out of six tasks, with Winogrande being the only exception.
Similarly , GPT-3.5 also shows improvements in most tasks, succeeding in four out of six. The exceptions are Winogrande and MathQA, with MathQA noted as the most challenging one across all tasks.
The LLaMA 2 models, spanning the 7B, 13B, and 70B variants, improve in half of the evaluated tasks. Their performance is consistent across model sizes. They excel in ARC Challenge, HellaSwag, and MMLU but face challenges in the other three tasks.

A noteworthy observation is that all models, regardless of their capability, exhibit reduced performance on the Winogrande task after applying our two-hop strategy. This includes both the high-end models like PaLM 2 and Gemini Pro, as well as the smaller scale LLaMA 2-7B model. The reasons for this decline are not immediately apparent, as factors such as model strength, task difficulty, and specific sensitivity to Winogrande do not directly explain the reduced effectiveness. Winogrande stands out due to its unique characteristics: it offers only two options per question and employs a cloze-test format rather than standard question-answering. We hypothesize that the limited number of options or the specific task type might alter the LLM's preference distribution, impacting the efficacy of our black-box strategy.

\subsection{Breakdown MMLU Subtasks}
To gain deeper insights, we conducted a detailed breakdown of the MMLU's 57 subtasks, examining closely how each method we proposed affects these subtasks. Figure \ref{fig:gray_box_improvement} offers a comprehensive view of how the MMLU's 57 subtasks respond to both the probability weighting and calibration methods within a gray-box scenario.
Consistent with the findings reported in Table \ref{graybox-results}, most of subtasks improve after applying our proposed methods.
Specifically, within the probability weighting analysis, declines are observed in only 1, 6, and 3 subtasks across the \textit{token}, \textit{order}, and \textit{both} settings, respectively.
This translates to an average of merely about 6\% of tasks not deriving benefits from the weighting method.
Upon closer examination, \textit{virology} emerges as the subtask experiencing a decline across all three sensitivity settings.
Among subtasks with notable decreases, \textit{machine learning} in the both setting shows a 3.57\% drop, while \textit{moral scenarios} and \textit{business ethics} in the order setting decline by 3.46\% and 3\%, respectively.

Regarding the probability calibration method, on average, more than 78\% of the subtasks improved with our approach, with over 30\% of them having at least a 1\% increase in accuracy. Recall that, in Table \ref{graybox-results}, the calibration method significantly outperforms the weighting method in the MathQA task. This trend extends across MMLU subtasks, with STEM-related tasks showing the most substantial gains. For instance, in the \textit{both} setting, the top beneficiaries include \textit{elementary mathematics}, \textit{high school mathematics}, \textit{college physics}, and \textit{college chemistry}, with improvements of 14.29\%, 12.22\%, 11.27\%, and 7.00\%, respectively, outshining other subtasks. This phenomenon is shown in the bottom three diagrams of Figure \ref{fig:gray_box_improvement}. Due to space constraints, detailed breakdowns of the gray-box method are presented in Appendix \ref{appendix-gray}, within Table \ref{mmlu-gray-box-weighting} and \ref{mmlu-gray-box-calibration}. Figure \ref{fig:black_box_acc_improvement_token}, \ref{fig:black_box_acc_improvement_order}, and \ref{fig:black_box_acc_improvement_both} in Appendix \ref{appendix-black} displays the results of the black-box two-hop strategy, with LLMs, where only generated text is accessible. Despite this limitation, more than half of the MMLU subtasks show improvement after our method's application. This enhancement is observed across various models, from the strongest to smaller ones like the 7B and 13B models.

\subsection{Cost Analysis}
Our method prioritizes cost-effectiveness by minimizing the need for numerous permutations or voting on costly chain-of-thought (CoT) candidates. For the probability weighting method, each question \(q\) needs two requests to calculate the weighted probability. In contrast, the probability calibration and gray-box methods require a validation set of approximately 200 samples to compute the distributions \(D_{forward}\) and \(D_{backward}\). Calibration of either \(r_{forward}\) or \(r_{backward}\) alone requires just one request per question; The black-box method also demands two requests per question \(q\).
Furthermore, the total expense for all experiments conducted in this study was under \$400 USD, covering six models and six benchmarks. Notably, the use of PaLM 2 and Gemini Pro was temporarily free. For specific costs associated with the OpenAI API and Azure AI Studio, please refer to the official documentations.

\section{Conclusion}
This study investigate into the effects of token and order sensitivity on LLMs when addressing selection problems, incorporating an empirical analysis of both powerful commercial LLMs such as Gemini Pro and GPT-3.5 and open-sources models like  LLaMA 2. By concentrating on zero-shot settings, we aim to isolate and better understand biases that previous works identified in in-context demonstrations, thereby offering a clearer perspective on how these sensitivities influence LLM decision-making processes. Our findings underscore the significance of task difficulty as a crucial determinant of sensitivity impact on LLM performance.

Moreover, we introduce cost-effective mitigation strategies, including gray-box and black-box approaches, tailored for practical application scenarios. The results demonstrate that our gray-box methods, namely probability weighting and probability calibration, outperform baselines with minimal additional expenditure, contrasting with more complex methods like majority voting. Additionally, our two-hop strategy for black-box scenarios proves to be effective in a significant portion of tasks. We anticipate that our contributions will aid future research in enhancing the robustness of LLMs across various types of selection problem applications.

\section{Limitation}
While this study contributes valuable insights into mitigating selection biases in LLMs, we acknowledge several limitations that warrant consideration.
Firstly, the gray-box strategies proposed for alleviating selection biases may encounter constraints when applied to certain black-box LLM APIs.
The efficacy of these strategies heavily relies on the availability of probability information, which may be restricted in externally hosted APIs.

Secondly, the exploration of mitigation strategies primarily focuses on the gray-box and black-box settings, leaving the examination of further mitigation strategies in white-box open-source models unexplored, and we recognize it as a potential avenue for future research.
Investigating mitigation strategies within white-box open-source models could provide a more comprehensive understanding of how selection biases manifest and can be addressed in models where internal workings are transparent.

\section*{Acknowledgements}
This work was supported by National Science and Technology Council, Taiwan, under grants MOST 110-2221-E-002-128-MY3, NSTC 112-2634-F-002-005 -, and Ministry of Education (MOE) in Taiwan, under grants NTU-113L900901.

\bibliography{main}

\begin{thebibliography}{29}
\providecommand{\natexlab}[1]{#1}

\bibitem[{Anil(2023{\natexlab{a}})}]{Anil2023GeminiAF}
Anil, R. (2023{\natexlab{a}}).
\newblock {Gemini: A Family of Highly Capable Multimodal Models}.
\newblock \emph{ArXiv}, abs/2312.11805.
\newblock URL: \url{https://api.semanticscholar.org/CorpusID:266361876}.

\bibitem[{Achiam(2023)}]{Achiam2023GPT4TR}
Achiam, R. (2023).
\newblock {GPT-4 Technical Report}.
\newblock \emph{ArXiv}.
\newblock URL: \url{https://api.semanticscholar.org/CorpusID:257532815}.

\bibitem[{Anil(2023{\natexlab{b}})}]{Anil2023PaLM2T}
Anil, R., Dai, A.~M., Firat, O., Johnson, M., Lepikhin, D., Passos, A.,
  Shakeri, S., Taropa, E., Bailey, P., Chen, Z., et~al. (2023{\natexlab{b}}).
\newblock {PaLM 2 technical report}.
\newblock \emph{arXiv preprint arXiv:2305.10403}.

\bibitem[{Touvron et~al.(2023)Touvron, Martin, Stone, Albert, Almahairi,
  Babaei, Bashlykov, Batra, Bhargava, Bhosale et~al.}]{touvron2023llama}
Touvron, H., Martin, L., Stone, K., Albert, P., Almahairi, A., Babaei, Y.,
  Bashlykov, N., Batra, S., Bhargava, P., Bhosale, S., et~al. (2023).
\newblock {Llama 2: Open Foundation and Fine-Tuned Chat Models}.
\newblock \emph{arXiv preprint arXiv:2307.09288}.

\bibitem[{Beeching et~al.(2023)Beeching, Fourrier, Habib, Han, Lambert, Rajani,
  Sanseviero, Tunstall, and Wolf}]{open-llm-leaderboard}
Beeching, E., Fourrier, C., Habib, N., Han, S., Lambert, N., Rajani, N.,
  Sanseviero, O., Tunstall, L., and Wolf, T. (2023).
\newblock {Open LLM Leaderboard}.
\newblock \emph{Hugging Face}.
\newblock URL: \url{https://huggingface.co/spaces/HuggingFaceH4/open_llm_leaderboard}.

\bibitem[{Gao et~al.(2023)Gao, Tow, Abbasi, Biderman, Black, DiPofi, Foster,
  Golding, Hsu, Le~Noac'h et~al.}]{eval-harness}
Gao, L., Tow, J., Abbasi, B., Biderman, S., Black, S., DiPofi, A., Foster, C.,
  Golding, L., Hsu, J., Le~Noac'h, A., et~al. (2023).
\newblock {A framework for few-shot language model evaluation}.
\newblock \emph{Zenodo}.
\newblock URL: \url{https://zenodo.org/records/10256836}.

\bibitem[{OpenAI(2022)}]{OpenAIChatGPT}
OpenAI (2022).
\newblock {Introducing ChatGPT}.
\newblock \emph{OpenAI}.
\newblock URL: \url{https://openai.com/blog/chatgpt}.

\bibitem[{Clark et~al.(2018)Clark, Cowhey, Etzioni, Khot, Sabharwal, Schoenick,
  and Tafjord}]{clark2018think}
Clark, P., Cowhey, I., Etzioni, O., Khot, T., Sabharwal, A., Schoenick, C., and
  Tafjord, O. (2018).
\newblock {Think you have Solved Question Answering? Try ARC, the AI2
  Reasoning Challenge}.
\newblock \emph{arXiv preprint arXiv:1803.05457}.

\bibitem[{Zellers et~al.(2019)Zellers, Holtzman, Bisk, Farhadi, and
  Choi}]{zellers2019hellaswag}
Zellers, R., Holtzman, A., Bisk, Y., Farhadi, A., and Choi, Y. (2019).
\newblock {HellaSwag: Can a Machine Really Finish Your Sentence?}
\newblock In \emph{Proceedings of the 57th Annual Meeting of the Association
  for Computational Linguistics}, pages 4791--4800. Association for
  Computational Linguistics.
\newblock URL: \url{https://aclanthology.org/P19-1472}.

\bibitem[{Hendrycks et~al.(2021)Hendrycks, Burns, Basart, Zou, Mazeika, Song,
  and Steinhardt}]{hendrycks2021measuring}
Hendrycks, D., Burns, C., Basart, S., Zou, A., Mazeika, M., Song, D., and
  Steinhardt, J. (2021).
\newblock {Measuring Massive Multitask Language Understanding}.
\newblock In \emph{International Conference on Learning Representations}.
\newblock URL: \url{https://openreview.net/forum?id=d7KBjmI3GmQ}.

\bibitem[{Sakaguchi et~al.(2019)Sakaguchi, Bras, Bhagavatula, and
  Choi}]{WINOGRANDE}
Sakaguchi, K., Bras, R.~L., Bhagavatula, C., and Choi, Y. (2019).
\newblock {WINOGRANDE: An Adversarial Winograd Schema Challenge at Scale}.
\newblock \emph{arXiv preprint arXiv:1907.10641}.

\bibitem[{Mihaylov et~al.(2018)Mihaylov, Clark, Khot, and
  Sabharwal}]{Mihaylov2018CanAS}
Mihaylov, T., Clark, P., Khot, T., and Sabharwal, A. (2018).
\newblock {Can a Suit of Armor Conduct Electricity? A New Dataset for Open Book
  Question Answering}.
\newblock In \emph{Proceedings of the 2018 Conference on Empirical Methods in
  Natural Language Processing}, pages 2381--2391. Association for Computational
  Linguistics.
\newblock URL: \url{https://aclanthology.org/D18-1260}.

\bibitem[{Amini et~al.(2019)Amini, Gabriel, Lin, Koncel-Kedziorski, Choi, and
  Hajishirzi}]{Amini2019MathQATI}
Amini, A., Gabriel, S., Lin, S., Koncel-Kedziorski, R., Choi, Y., and
  Hajishirzi, H. (2019).
\newblock {MathQA: Towards Interpretable Math Word Problem Solving with
  Operation-Based Formalisms}.
\newblock \emph{ArXiv}, abs/1905.13319.
\newblock URL: \url{https://api.semanticscholar.org/CorpusID:173188048}.

\bibitem[{Zheng et~al.(2023{\natexlab{a}})Zheng, Chiang, Sheng, Zhuang, Wu,
  Zhuang, Lin, Li, Li, Xing et~al.}]{Zheng2023JudgingLW}
Zheng, L., Chiang, W.-L., Sheng, Y., Zhuang, S., Wu, Z., Zhuang, Y., Lin, Z.,
  Li, Z., Li, D., Xing, E., et~al. (2023{\natexlab{a}}).
\newblock {Judging LLM-as-a-Judge with MT-Bench and Chatbot Arena}.
\newblock In \emph{Advances in Neural Information Processing Systems}, pages
  46595--46623. A. Oh, T. Naumann, A. Globerson, K. Saenko, M. Hardt, S.
  Levine (editors).
\newblock URL: \url{https://proceedings.neurips.cc/paper_files/paper/2023/file/91f18a1287b398d378ef22505bf41832-Paper-Datasets_and_Benchmarks.pdf}.

\bibitem[{Wang et~al.(2023{\natexlab{a}})Wang, Li, Chen, Zhu, Lin, Cao, Liu,
  Liu, and Sui}]{Wang2023LargeLM}
Wang, P., Li, L., Chen, L., Zhu, D., Lin, B., Cao, Y., Liu, Q., Liu, T., and
  Sui, Z. (2023{\natexlab{a}}).
\newblock {Large Language Models are not Fair Evaluators}.
\newblock \emph{ArXiv}, abs/2305.17926.
\newblock URL: \url{https://api.semanticscholar.org/CorpusID:258960339}.

\bibitem[{Zheng et~al.(2023{\natexlab{b}})Zheng, Zhou, Meng, Zhou, and
  Huang}]{Zheng2023LargeLM}
Zheng, C., Zhou, H., Meng, F., Zhou, J., and Huang, M. (2023{\natexlab{b}}).
\newblock {Large Language Models Are Not Robust Multiple Choice Selectors}.
\newblock In \emph{The Twelfth International Conference on Learning
  Representations}.
\newblock URL: \url{https://openreview.net/forum?id=shr9PXz7T0}.

\bibitem[{Pezeshkpour and Hruschka(2023)}]{Pezeshkpour2023LargeLM}
Pezeshkpour, P. and Hruschka, E. (2023).
\newblock {Large Language Models Sensitivity to The Order of Options in
  Multiple-Choice Questions}.
\newblock \emph{ArXiv}, abs/2308.11483.
\newblock URL: \url{https://api.semanticscholar.org/CorpusID:261064970}.

\bibitem[{Shen et~al.(2023)Shen, Cheng, Nguyen, You, and Bing}]{Shen2023LargeLM}
Shen, C., Cheng, L., Nguyen, X.-P., You, Y., and Bing, L. (2023).
\newblock {Large Language Models are Not Yet Human-Level Evaluators for
  Abstractive Summarization}.
\newblock In \emph{Findings of the Association for Computational Linguistics:
  EMNLP 2023}, pages 4215--4233. Association for Computational Linguistics.
\newblock URL: \url{https://aclanthology.org/2023.findings-emnlp.278}.

\bibitem[{Chiang and Lee(2023)}]{Chiang2023CanLL}
Chiang, C.-H. and Lee, H.-Y. (2023).
\newblock {Can Large Language Models Be an Alternative to Human Evaluations?}
\newblock In \emph{Annual Meeting of the Association for Computational
  Linguistics}.
\newblock URL: \url{https://api.semanticscholar.org/CorpusID:258461287}.

\bibitem[{Chan et~al.(2023)Chan, Chen, Su, Yu, Xue, Zhang, Fu, and
  Liu}]{Chan2023ChatEvalTB}
Chan, C.-M., Chen, W., Su, Y., Yu, J., Xue, W., Zhang, S., Fu, J., and Liu, Z.
  (2023).
\newblock {ChatEval: Towards Better LLM-based Evaluators through Multi-Agent
  Debate}.
\newblock \emph{ArXiv}, abs/2308.07201.
\newblock URL: \url{https://api.semanticscholar.org/CorpusID:260887105}.

\bibitem[{Si et~al.(2023)Si, Friedman, Joshi, Feng, Chen, and
  He}]{Si2023MeasuringIB}
Si, C., Friedman, D., Joshi, N., Feng, S., Chen, D., and He, H. (2023).
\newblock {Measuring Inductive Biases of In-Context Learning with
  Underspecified Demonstrations}.
\newblock In \emph{Proceedings of the 61st Annual Meeting of the Association
  for Computational Linguistics (Volume 1: Long Papers)}, pages 11289--11310.
  Association for Computational Linguistics.
\newblock URL: \url{https://aclanthology.org/2023.acl-long.632}.

\bibitem[{Zhao et~al.(2021)Zhao, Wallace, Feng, Klein, and
  Singh}]{Zhao2021CalibrateBU}
Zhao, T., Wallace, E., Feng, S., Klein, D., and Singh, S. (2021).
\newblock {Calibrate Before Use: Improving Few-Shot Performance of Language
  Models}.
\newblock In \emph{International Conference on Machine Learning}.
\newblock URL: \url{https://api.semanticscholar.org/CorpusID:231979430}.

\bibitem[{Fei et~al.(2023)Fei, Hou, Chen, and Bosselut}]{Fei2023MitigatingLB}
Fei, Y., Hou, Y., Chen, Z., and Bosselut, A. (2023).
\newblock {Mitigating Label Biases for In-context Learning}.
\newblock In \emph{Proceedings of the 61st Annual Meeting of the Association
  for Computational Linguistics (Volume 1: Long Papers)}, pages 14014--14031.
  Association for Computational Linguistics.
\newblock URL: \url{https://aclanthology.org/2023.acl-long.783}.

\bibitem[{Wang et~al.(2023{\natexlab{b}})Wang, Liu, Park, Chen, and
  Xiao}]{Wang2023AdversarialDA}
Wang, J., Liu, Z.-Y., Park, K.~H., Chen, M., and Xiao, C. (2023{\natexlab{b}}).
\newblock {Adversarial Demonstration Attacks on Large Language Models}.
\newblock \emph{ArXiv}, abs/2305.14950.
\newblock URL: \url{https://api.semanticscholar.org/CorpusID:258865399}.

\bibitem[{Zhu et~al.(2023)Zhu, Wang, Zhou, Wang, Chen, Wang, Yang, Ye, Gong,
  Zhang, and Xie}]{Zhu2023PromptBenchTE}
Zhu, K., Wang, J., Zhou, J., Wang, Z., Chen, H., Wang, Y., Yang, L., Ye, W.,
  Gong, N.~Z., Zhang, Y., and Xie, X. (2023).
\newblock {PromptBench: Towards Evaluating the Robustness of Large Language
  Models on Adversarial Prompts}.
\newblock \emph{ArXiv}, abs/2306.04528.
\newblock URL: \url{https://api.semanticscholar.org/CorpusID:259095572}.

\bibitem[{Robinson and Wingate(2023)}]{Robinson2022LeveragingLL}
Robinson, J. and Wingate, D. (2023).
\newblock {Leveraging Large Language Models for Multiple Choice Question
  Answering}.
\newblock In \emph{The Eleventh International Conference on Learning
  Representations}.
\newblock URL: \url{https://openreview.net/forum?id=yKbprarjc5B}.

\end{thebibliography}

\appendix

\section{Data statistics details}
\label{data-details-appendix}
Table \ref{tab:benchmark} shows the data statistics details across six tasks: ARC-Challenge, HellaSwag, MMLU, Winogrande, MathQA, and OpenBookQA.

\begin{table}[ht]
\centering
\small
\begin{tabular}{lcc}
\hline
\textbf{Tasks}         & \textbf{\# Samples} & \textbf{\# Options} \\ \hline
Winogrande    &      1,267        &      2      \\ \hline
ARC-Challenge &      1,165        &      4      \\
HellaSwag     &      10,042        &      4      \\
MMLU          &      14,042        &      4      \\
OpenBookQA    &      500        &      4      \\ \hline
MathQA        &      2,985        &      5      \\ \hline
\end{tabular}
\caption{Data statistics of our benchmarks.}
\label{tab:benchmark}
\end{table}

\section{Prompt templates}
\label{sec:prompt_format}
We list all the prompt templates used in our experiments, including three different sensitivity settings: \textit{token}, \textit{order}, and \textit{both}. These templates are presented in Figures \ref{fig:prompt_example}, \ref{fig:prompt_example2}, and \ref{fig:prompt_example3}, corresponding to each setting respectively.

\section{Detailed Sensitivity Experiment Results}
\label{detailed-sensitivity-result}
Tables \ref{mmlu-palm2} through \ref{mmlu-llama-70b} provide detailed experimental results for the MMLU, covering its 57 subtasks for each of the following models: PaLM 2, Gemini Pro, GPT-3.5, LLaMA 2 7B, LLaMA 2 13B, and LLaMA 2 70B, respectively.

\section{LLMs' Option proportion statistics}
\label{llm-option-stats}
Table \ref{option-portion-first} through \ref{option-portion-last} provide detailed information on the tendency of each option and the ground truth label proportion across five datasets: HellaSwag, MMLU, Winogrande, MathQA, and OpenBookQA.

\section{Gray-Box Results of MMLU Subtasks}
\label{appendix-gray}
Tables \ref{mmlu-gray-box-weighting} and \ref{mmlu-gray-box-calibration} detail the results for the MMLU's 57 subtasks following the application of our gray-box strategies, including probability weighting and calibration.

\section{Black-Box Results of MMLU Subtasks}
\label{appendix-black}
\begin{figure*}[h]
\definecolor{darkgreen}{rgb}{0.0, 0.5, 0.0}
\centering
\includegraphics[width=\linewidth]{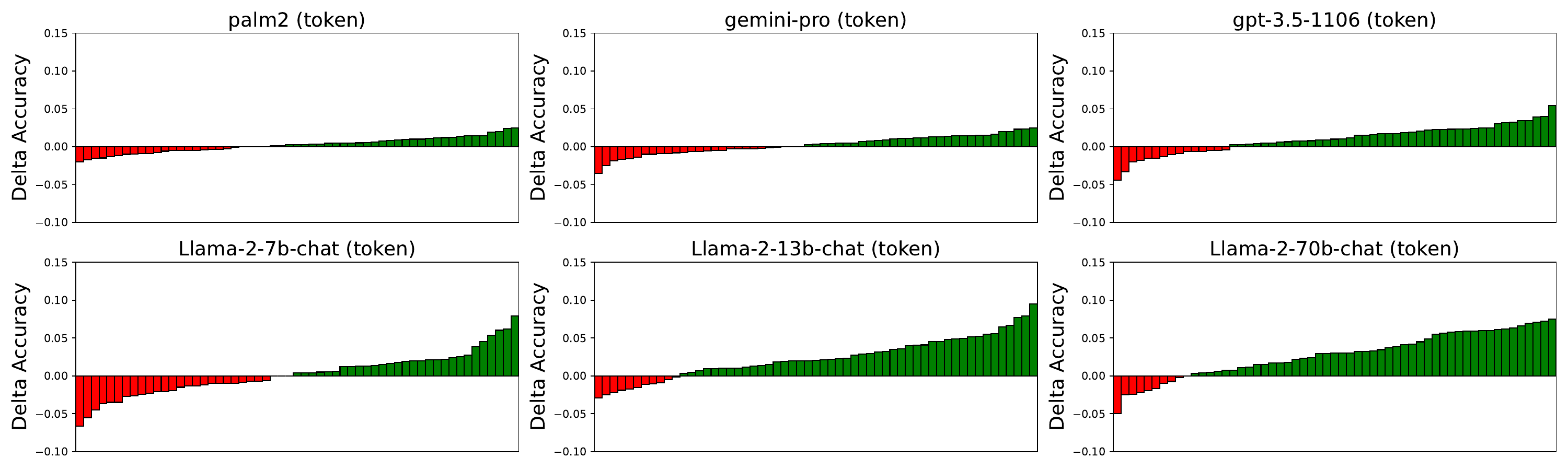}
\caption{
Accuracy Difference Distribution Across 57 MMLU Subtasks in the Black-Box Scenario With Token Sensitivity Setting: Subtasks are sorted by the difference in accuracy from low to high, indicating that subtasks towards the right benefit more from our methodology. Improvements are marked in \textcolor{darkgreen}{green}, whereas declines in performance are highlighted in \textcolor{red} {red}.
}
\label{fig:black_box_acc_improvement_token}
\end{figure*}

\begin{figure*}[h]
\centering
\definecolor{darkgreen}{rgb}{0.0, 0.5, 0.0}
\includegraphics[width=\linewidth]{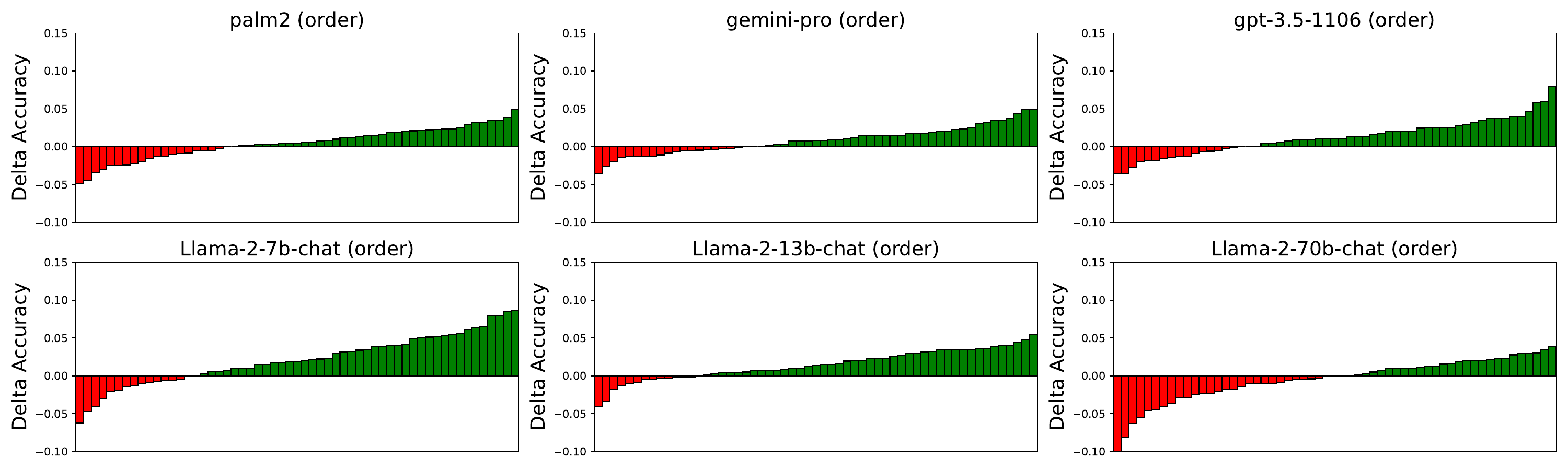}
\caption{
Accuracy Difference Distribution Across 57 MMLU Subtasks in the Black-Box Scenario With Order Sensitivity Setting: Subtasks are sorted by the difference in accuracy from low to high, indicating that subtasks towards the right benefit more from our methodology. Improvements are marked in \textcolor{darkgreen}{green}, whereas declines in performance are highlighted in \textcolor{red} {red}.
}
\label{fig:black_box_acc_improvement_order}
\end{figure*}

\begin{figure*}[t]
\definecolor{darkgreen}{rgb}{0.0, 0.5, 0.0}
\centering
\includegraphics[width=\linewidth]{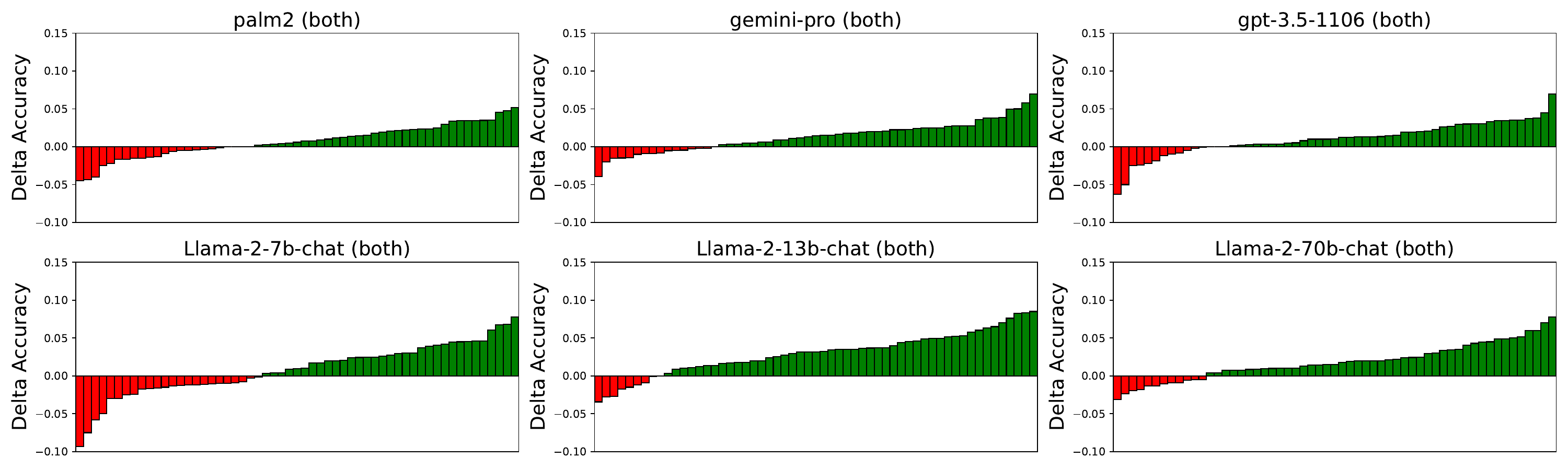}
\caption{
Accuracy Difference Distribution Across 57 MMLU Subtasks in the Black-Box Scenario: Subtasks are sorted by the difference in accuracy from low to high, indicating that subtasks towards the right benefit more from our methodology. Improvements are marked in \textcolor{darkgreen}{green}, whereas declines in performance are highlighted in \textcolor{red} {red}.
}
\label{fig:black_box_acc_improvement_both}
\end{figure*}

Figure \ref{fig:black_box_acc_improvement_token}, \ref{fig:black_box_acc_improvement_order} and \ref{fig:black_box_acc_improvement_both} show the distribution of accuracy differences resulting from the black-box approach, specifically within the \textit{token}, \textit{order} and \textit{both} sensitivity settings.

\begin{figure*}
\centering
\small
\begin{tabular}{l}
\hline
[System] \\
Please carefully read the following questions and choices. Select the most suitable one. Output your final verdict \\
by strictly following this prompt: Indicate your choice by placing it inside double square brackets, with a single\\
character representing the chosen option.For example, [[<single\_character>]].\\\\

[The start of question]\\
\textcolor{orange}{As of 2020, which architecture is best for classifying high-resolution images?} \\
{[The end of question]} \\\\
{[The start of choice \textcolor{blue}{A}]}\\
\textcolor{orange}{convolutional networks}\\
{[The end of choice \textcolor{blue}{A}]} \\\\
{[The start of choice \textcolor{blue}{B}]}\\
\textcolor{orange}{graph networks}\\
{[The end of choice \textcolor{blue}{B}]}\\\\
{[The start of choice \textcolor{blue}{C}]}\\
\textcolor{orange}{fully connected networks}\\
{[The end of choice \textcolor{blue}{C}]} \\\\
{[The start of choice \textcolor{blue}{D}]}\\
\textcolor{orange}{RBF networks}\\
{[The end of choice \textcolor{blue}{D}]}\\
\hline
[System] \\
Please carefully read the following questions and choices. Select the most suitable one. Output your final verdict \\
by strictly following this prompt: Indicate your choice by placing it inside double square brackets, with a single\\
character representing the chosen option.For example, [[<single\_character>]].\\\\

[The start of question]\\
\textcolor{orange}{As of 2020, which architecture is best for classifying high-resolution images?} \\
{[The end of question]} \\\\
{[The start of choice \textcolor{blue}{D}]}\\
\textcolor{orange}{convolutional networks}\\
{[The end of choice \textcolor{blue}{D}]} \\\\
{[The start of choice \textcolor{blue}{C}]}\\
\textcolor{orange}{graph networks}\\
{[The end of choice \textcolor{blue}{C}]}\\\\
{[The start of choice \textcolor{blue}{B}]}\\
\textcolor{orange}{fully connected networks}\\
{[The end of choice \textcolor{blue}{B}]} \\\\
{[The start of choice \textcolor{blue}{A}]}\\
\textcolor{orange}{RBF networks}\\
{[The end of choice \textcolor{blue}{A}]}\\
\hline
\end{tabular}
\caption{
Prompt template illustrating the \textit{token} sensitivity setting for each question \(q\). The upper part represents \(r_{forward}\), and the lower part corresponds to \(r_{backward}\). Option symbols are highlighted in \textcolor{blue}{blue}, while both the question text and option contents are highlighted in \textcolor{orange}{orange}. Other text shown in black remains consistent across questions.
}
\label{fig:prompt_example}
\end{figure*}

\begin{figure*}
\centering
\small
\begin{tabular}{l}
\hline
[System] \\
Please carefully read the following questions and choices. Select the most suitable one. Output your final verdict \\
by strictly following this prompt: Indicate your choice by placing it inside double square brackets, with a single\\
character representing the chosen option.For example, [[<single\_character>]].\\\\

[The start of question]\\
\textcolor{orange}{As of 2020, which architecture is best for classifying high-resolution images?} \\
{[The end of question]} \\\\
{[The start of choice \textcolor{blue}{A}]}\\
\textcolor{orange}{convolutional networks}\\
{[The end of choice \textcolor{blue}{A}]} \\\\
{[The start of choice \textcolor{blue}{B}]}\\
\textcolor{orange}{graph networks}\\
{[The end of choice \textcolor{blue}{B}]}\\\\
{[The start of choice \textcolor{blue}{C}]}\\
\textcolor{orange}{fully connected networks}\\
{[The end of choice \textcolor{blue}{C}]} \\\\
{[The start of choice \textcolor{blue}{D}]}\\
\textcolor{orange}{RBF networks}\\
{[The end of choice \textcolor{blue}{D}]}\\
\hline
[System] \\
Please carefully read the following questions and choices. Select the most suitable one. Output your final verdict \\
by strictly following this prompt: Indicate your choice by placing it inside double square brackets, with a single\\
character representing the chosen option.For example, [[<single\_character>]].\\\\

[The start of question]\\
\textcolor{orange}{As of 2020, which architecture is best for classifying high-resolution images?} \\
{[The end of question]} \\\\
{[The start of choice \textcolor{blue}{D}]}\\
\textcolor{orange}{RBF networks}\\
{[The end of choice \textcolor{blue}{D}]}\\\\
{[The start of choice \textcolor{blue}{C}]}\\
\textcolor{orange}{fully connected networks}\\
{[The end of choice \textcolor{blue}{C}]} \\\\
{[The start of choice \textcolor{blue}{B}]}\\
\textcolor{orange}{graph networks}\\
{[The end of choice \textcolor{blue}{B}]}\\\\
{[The start of choice \textcolor{blue}{A}]}\\
\textcolor{orange}{convolutional networks}\\
{[The end of choice \textcolor{blue}{A}]} \\
\hline
\end{tabular}
\caption{
Prompt template illustrating the \textit{order} sensitivity setting for each question \(q\). The upper part represents \(r_{forward}\), and the lower part corresponds to \(r_{backward}\). Option symbols are highlighted in \textcolor{blue}{blue}, while both the question text and option contents are highlighted in \textcolor{orange}{orange}. Other text shown in black remains consistent across questions.
}
\label{fig:prompt_example2}
\end{figure*}

\begin{figure*}
\centering
\small
\begin{tabular}{l}
\hline
[System] \\
Please carefully read the following questions and choices. Select the most suitable one. Output your final verdict \\
by strictly following this prompt: Indicate your choice by placing it inside double square brackets, with a single\\
character representing the chosen option.For example, [[<single\_character>]].\\\\

[The start of question]\\
\textcolor{orange}{As of 2020, which architecture is best for classifying high-resolution images?} \\
{[The end of question]} \\\\
{[The start of choice \textcolor{blue}{A}]}\\
\textcolor{orange}{convolutional networks}\\
{[The end of choice \textcolor{blue}{A}]} \\\\
{[The start of choice \textcolor{blue}{B}]}\\
\textcolor{orange}{graph networks}\\
{[The end of choice \textcolor{blue}{B}]}\\\\
{[The start of choice \textcolor{blue}{C}]}\\
\textcolor{orange}{fully connected networks}\\
{[The end of choice \textcolor{blue}{C}]} \\\\
{[The start of choice \textcolor{blue}{D}]}\\
\textcolor{orange}{RBF networks}\\
{[The end of choice \textcolor{blue}{D}]}\\
\hline
[System] \\
Please carefully read the following questions and choices. Select the most suitable one. Output your final verdict \\
by strictly following this prompt: Indicate your choice by placing it inside double square brackets, with a single\\
character representing the chosen option.For example, [[<single\_character>]].\\\\

[The start of question]\\
\textcolor{orange}{As of 2020, which architecture is best for classifying high-resolution images?} \\
{[The end of question]} \\\\
{[The start of choice \textcolor{blue}{A}]}\\
\textcolor{orange}{RBF networks}\\
{[The end of choice \textcolor{blue}{A}]} \\\\
{[The start of choice \textcolor{blue}{B}]}\\
\textcolor{orange}{fully connected networks}\\
{[The end of choice \textcolor{blue}{B}]}\\\\
{[The start of choice \textcolor{blue}{C}]}\\
\textcolor{orange}{graph networks}\\
{[The end of choice \textcolor{blue}{C}]} \\\\
{[The start of choice \textcolor{blue}{D}]}\\
\textcolor{orange}{convolutional networks}\\
{[The end of choice \textcolor{blue}{D}]}\\
\hline
\end{tabular}
\caption{
Prompt template illustrating the \textit{both} sensitivity setting for each question \(q\). The upper part represents \(r_{forward}\), and the lower part corresponds to \(r_{backward}\). Option symbols are highlighted in \textcolor{blue}{blue}, while both the question text and option contents are highlighted in \textcolor{orange}{orange}. Other text shown in black remains consistent across questions.}
\label{fig:prompt_example3}
\end{figure*}

\begin{table*}
\centering
\small
\begin{tabular}{l|c|cr|cr|cr}
\hline
 &  & \multicolumn{2}{c|}{\textbf{Token}} & \multicolumn{2}{c|}{\textbf{Order}} & \multicolumn{2}{c}{\textbf{Both}} \\
\multicolumn{1}{c|}{\textbf{Subtask}} & \textbf{\#sample} & \textbf{Avg.} & \textbf{Fluct.} & \textbf{Avg.} & \textbf{Fluct.} & \textbf{Avg.} & \textbf{Fluct.} \\
 & & \textbf{Acc} & \textbf{Rate} & \textbf{Acc} & \textbf{Rate} & \textbf{Acc} & \textbf{Rate} \\
\hline
abstract\_algebra & 100 & 32.0 & 15.0 & 32.5 & 47.0 & 33.5 & 34.0 \\
anatomy & 135 & 64.4 & 11.1 & 62.9 & 26.7 & 62.2 & 29.6 \\
astronomy & 152 & 73.7 & 12.5 & 74.3 & 28.9 & 74.3 & 28.3 \\
business\_ethics & 100 & 75.0 & 9.0 & 71.0 & 27.0 & 72.5 & 26.0 \\
clinical\_knowledge & 265 & 72.6 & 8.3 & 71.7 & 23.4 & 71.3 & 25.7 \\
\hline
college\_biology & 144 & 81.6 & 5.6 & 81.3 & 16.0 & 79.9 & 20.1 \\
college\_chemistry & 100 & 53.0 & 17.0 & 51.5 & 40.0 & 50.0 & 44.0 \\
college\_computer\_science & 100 & 44.0 & 20.0 & 44.0 & 38.0 & 42.0 & 46.0 \\
college\_mathematics & 100 & 32.0 & 27.0 & 32.0 & 41.0 & 29.5 & 54.0 \\
college\_medicine & 173 & 57.5 & 11.0 & 58.1 & 28.3 & 57.8 & 33.5 \\
\hline
college\_physics & 102 & 48.5 & 19.6 & 44.6 & 40.2 & 44.6 & 39.2 \\
computer\_security & 100 & 74.5 & 7.0 & 76.5 & 14.0 & 75.5 & 21.0 \\
conceptual\_physics & 235 & 63.8 & 12.8 & 60.2 & 26.8 & 59.6 & 31.9 \\
econometrics & 114 & 48.2 & 23.7 & 47.4 & 37.7 & 50.9 & 41.2 \\
electrical\_engineering & 145 & 61.0 & 11.0 & 61.7 & 37.2 & 57.2 & 42.1 \\
\hline
elementary\_mathematics & 378 & 44.3 & 21.2 & 45.8 & 52.6 & 44.4 & 55.8 \\
formal\_logic & 126 & 51.2 & 31.7 & 49.6 & 42.9 & 46.8 & 46.8 \\
global\_facts & 100 & 34.5 & 26.0 & 40.5 & 50.0 & 38.5 & 56.0 \\
high\_school\_biology & 310 & 81.0 & 7.1 & 78.7 & 15.5 & 79.7 & 17.4 \\
high\_school\_chemistry & 203 & 56.9 & 17.7 & 54.2 & 36.0 & 54.4 & 40.9 \\
\hline
high\_school\_computer\_science & 100 & 69.5 & 16.0 & 72.0 & 25.0 & 70.5 & 24.0 \\
high\_school\_european\_history & 165 & 77.6 & 6.1 & 76.4 & 15.8 & 78.5 & 15.8 \\
high\_school\_geography & 198 & 82.8 & 8.1 & 81.8 & 12.6 & 80.8 & 14.6 \\
high\_school\_government\_and\_politics & 193 & 90.7 & 5.2 & 90.4 & 7.8 & 90.4 & 7.8 \\
high\_school\_macroeconomics & 390 & 65.1 & 11.5 & 64.0 & 24.6 & 63.6 & 27.2 \\
\hline
high\_school\_mathematics & 270 & 31.1 & 31.5 & 30.9 & 51.5 & 30.6 & 63.3 \\
high\_school\_microeconomics & 238 & 76.5 & 8.4 & 76.1 & 18.5 & 75.8 & 19.7 \\
high\_school\_physics & 151 & 36.8 & 19.2 & 37.7 & 41.7 & 37.4 & 42.4 \\
high\_school\_psychology & 545 & 88.6 & 3.7 & 86.8 & 11.6 & 87.4 & 11.7 \\
high\_school\_statistics & 216 & 58.3 & 15.7 & 54.2 & 38.9 & 52.1 & 42.1 \\
\hline
high\_school\_us\_history & 204 & 80.9 & 7.4 & 80.4 & 18.1 & 81.1 & 18.6 \\
high\_school\_world\_history & 237 & 83.3 & 3.4 & 82.7 & 11.8 & 85.2 & 13.5 \\
human\_aging & 223 & 69.7 & 9.0 & 70.4 & 19.3 & 68.4 & 22.0 \\
human\_sexuality & 131 & 75.2 & 9.9 & 75.9 & 23.7 & 74.8 & 29.0 \\
international\_law & 121 & 80.2 & 3.3 & 81.0 & 19.8 & 79.3 & 21.5 \\
\hline
jurisprudence & 108 & 81.0 & 3.7 & 80.1 & 17.6 & 81.0 & 13.0 \\
logical\_fallacies & 163 & 81.9 & 4.3 & 82.5 & 13.5 & 81.3 & 16.0 \\
machine\_learning & 112 & 42.4 & 17.0 & 44.2 & 31.3 & 44.2 & 25.9 \\
management & 103 & 86.4 & 8.7 & 85.0 & 11.7 & 85.4 & 13.6 \\
marketing & 234 & 86.1 & 2.6 & 87.4 & 9.4 & 86.5 & 11.5 \\
\hline
medical\_genetics & 100 & 69.0 & 16.0 & 68.5 & 20.0 & 70.5 & 26.0 \\
miscellaneous & 783 & 82.0 & 6.9 & 82.4 & 13.0 & 82.2 & 13.4 \\
moral\_disputes & 346 & 70.4 & 9.8 & 70.1 & 21.7 & 69.7 & 22.5 \\
moral\_scenarios & 895 & 32.1 & 79.6 & 31.3 & 16.2 & 34.9 & 76.1 \\
nutrition & 306 & 67.0 & 9.5 & 66.8 & 24.8 & 67.2 & 25.8 \\
\hline
philosophy & 311 & 70.6 & 7.7 & 69.0 & 24.1 & 69.1 & 23.5 \\
prehistory & 324 & 75.8 & 10.5 & 75.9 & 20.4 & 75.3 & 24.1 \\
professional\_accounting & 282 & 50.9 & 18.4 & 51.6 & 34.8 & 50.0 & 39.4 \\
professional\_law & 1534 & 46.7 & 17.7 & 45.6 & 41.9 & 44.4 & 49.1 \\
professional\_medicine & 272 & 69.3 & 11.4 & 66.7 & 28.7 & 65.4 & 33.1 \\
\hline
professional\_psychology & 612 & 70.8 & 8.7 & 69.4 & 20.8 & 69.0 & 24.3 \\
public\_relations & 110 & 71.4 & 7.3 & 68.6 & 26.4 & 68.2 & 28.2 \\
security\_studies & 245 & 75.9 & 4.5 & 75.1 & 25.3 & 76.7 & 26.1 \\
sociology & 201 & 87.3 & 4.5 & 84.6 & 15.4 & 83.1 & 16.9 \\
us\_foreign\_policy & 100 & 86.5 & 9.0 & 86.5 & 11.0 & 83.5 & 16.0 \\
\hline
virology & 166 & 49.4 & 10.8 & 51.5 & 19.9 & 50.0 & 27.1 \\
world\_religions & 171 & 82.5 & 2.9 & 83.6 & 11.7 & 83.0 & 11.7 \\
\hline
\end{tabular}
\caption{\label{mmlu-palm2}
Results of the sensitivity experiment across 57 MMLU subtasks for PaLM 2, including different sensitivity settings: \textit{Token}, \textit{Order}, and \textit{Both}. \textbf{Avg. Acc} represents the mean of \(r_{forward}\) and \(r_{backward}\) accuracies for each setting.
}
\end{table*}

\begin{table*}
\centering
\small
\begin{tabular}{l|c|cr|cr|cr}
\hline
 &  & \multicolumn{2}{c|}{\textbf{Token}} & \multicolumn{2}{c|}{\textbf{Order}} & \multicolumn{2}{c}{\textbf{Both}} \\
\multicolumn{1}{c|}{\textbf{Subtask}} & \textbf{\#sample} & \textbf{Avg.} & \textbf{Fluct.} & \textbf{Avg.} & \textbf{Fluct.} & \textbf{Avg.} & \textbf{Fluct.} \\
 & & \textbf{Acc} & \textbf{Rate} & \textbf{Acc} & \textbf{Rate} & \textbf{Acc} & \textbf{Rate} \\
\hline
abstract\_algebra & 100 & 35.0 & 51.0 & 33.0 & 73.0 & 32.5 & 78.0 \\
anatomy & 135 & 60.0 & 18.5 & 61.1 & 34.1 & 60.0 & 31.1 \\
astronomy & 152 & 76.9 & 12.5 & 75.7 & 21.7 & 74.3 & 26.9 \\
business\_ethics & 100 & 62.5 & 14.0 & 64.5 & 38.0 & 65.5 & 38.0 \\
clinical\_knowledge & 265 & 74.3 & 12.8 & 74.2 & 23.4 & 72.8 & 27.9 \\
\hline
college\_biology & 144 & 81.3 & 11.8 & 81.9 & 20.1 & 81.3 & 22.9 \\
college\_chemistry & 100 & 46.5 & 28.0 & 49.0 & 39.0 & 47.5 & 56.0 \\
college\_computer\_science & 100 & 52.5 & 26.0 & 51.5 & 48.0 & 47.5 & 52.0 \\
college\_mathematics & 100 & 33.5 & 39.0 & 36.5 & 52.0 & 37.0 & 64.0 \\
college\_medicine & 173 & 67.9 & 12.1 & 66.2 & 31.2 & 63.9 & 32.9 \\
\hline
college\_physics & 102 & 37.7 & 34.3 & 41.2 & 43.1 & 41.7 & 43.1 \\
computer\_security & 100 & 72.5 & 10.0 & 73.5 & 21.0 & 74.5 & 21.0 \\
conceptual\_physics & 235 & 60.2 & 14.9 & 57.4 & 31.9 & 59.4 & 36.2 \\
econometrics & 114 & 44.3 & 32.5 & 45.2 & 48.2 & 45.2 & 50.9 \\
electrical\_engineering & 145 & 63.4 & 20.7 & 61.0 & 40.7 & 61.7 & 42.8 \\
\hline
elementary\_mathematics & 378 & 45.8 & 27.0 & 46.0 & 53.2 & 45.8 & 54.8 \\
formal\_logic & 126 & 47.2 & 17.5 & 47.6 & 46.8 & 47.2 & 51.6 \\
global\_facts & 100 & 43.5 & 40.0 & 39.5 & 78.0 & 40.0 & 66.0 \\
high\_school\_biology & 310 & 82.1 & 10.6 & 82.7 & 16.1 & 81.1 & 19.0 \\
high\_school\_chemistry & 203 & 55.7 & 25.6 & 53.4 & 38.9 & 51.7 & 44.3 \\
\hline
high\_school\_computer\_science & 100 & 76.0 & 13.0 & 73.5 & 26.0 & 75.5 & 26.0 \\
high\_school\_european\_history & 165 & 80.9 & 10.9 & 79.1 & 15.8 & 79.7 & 17.0 \\
high\_school\_geography & 198 & 78.8 & 13.1 & 83.3 & 15.7 & 79.8 & 22.2 \\
high\_school\_government\_and\_politics & 193 & 88.3 & 8.3 & 89.9 & 10.4 & 89.1 & 13.5 \\
high\_school\_macroeconomics & 390 & 66.9 & 10.8 & 66.3 & 30.0 & 65.3 & 31.5 \\
\hline
high\_school\_mathematics & 270 & 34.4 & 33.3 & 35.6 & 55.6 & 33.9 & 66.3 \\
high\_school\_microeconomics & 238 & 76.3 & 15.1 & 75.8 & 21.4 & 75.6 & 27.3 \\
high\_school\_physics & 151 & 38.7 & 33.1 & 40.4 & 48.3 & 40.4 & 53.6 \\
high\_school\_psychology & 545 & 85.6 & 7.2 & 85.7 & 10.8 & 85.3 & 14.3 \\
high\_school\_statistics & 216 & 59.3 & 20.8 & 56.7 & 41.7 & 53.0 & 47.7 \\
\hline
high\_school\_us\_history & 204 & 83.6 & 10.3 & 83.8 & 13.2 & 83.8 & 16.2 \\
high\_school\_world\_history & 237 & 86.3 & 8.4 & 84.2 & 16.0 & 83.8 & 18.6 \\
human\_aging & 223 & 71.3 & 9.9 & 69.3 & 25.1 & 69.5 & 29.6 \\
human\_sexuality & 131 & 77.5 & 14.5 & 74.0 & 20.6 & 74.8 & 26.7 \\
international\_law & 121 & 78.5 & 14.0 & 80.2 & 21.5 & 80.2 & 24.0 \\
\hline
jurisprudence & 108 & 78.7 & 13.9 & 78.7 & 18.5 & 75.0 & 22.2 \\
logical\_fallacies & 163 & 79.1 & 12.3 & 76.4 & 22.7 & 77.0 & 19.0 \\
machine\_learning & 112 & 41.9 & 27.7 & 43.3 & 52.7 & 47.3 & 58.0 \\
management & 103 & 83.0 & 6.8 & 80.1 & 17.5 & 80.6 & 20.4 \\
marketing & 234 & 88.0 & 6.0 & 88.7 & 12.4 & 88.2 & 13.7 \\
\hline
medical\_genetics & 100 & 75.0 & 12.0 & 73.0 & 22.0 & 72.5 & 25.0 \\
miscellaneous & 783 & 83.4 & 7.3 & 83.5 & 13.2 & 84.0 & 15.5 \\
moral\_disputes & 346 & 67.8 & 14.7 & 63.7 & 33.8 & 66.8 & 34.4 \\
moral\_scenarios & 895 & 40.9 & 51.4 & 35.6 & 86.5 & 36.3 & 90.3 \\
nutrition & 306 & 73.5 & 14.4 & 74.7 & 22.9 & 73.0 & 26.8 \\
\hline
philosophy & 311 & 74.3 & 13.5 & 72.0 & 27.7 & 70.6 & 28.9 \\
prehistory & 324 & 75.9 & 12.0 & 74.7 & 17.6 & 75.0 & 21.3 \\
professional\_accounting & 282 & 51.6 & 27.0 & 50.0 & 42.2 & 50.0 & 45.7 \\
professional\_law & 1534 & 51.1 & 26.5 & 50.1 & 37.5 & 49.4 & 50.1 \\
professional\_medicine & 272 & 75.0 & 12.9 & 73.7 & 23.5 & 69.1 & 30.9 \\
\hline
professional\_psychology & 612 & 69.9 & 15.4 & 68.5 & 25.0 & 68.6 & 25.8 \\
public\_relations & 110 & 66.8 & 16.4 & 65.5 & 30.9 & 63.6 & 30.0 \\
security\_studies & 245 & 73.9 & 16.7 & 73.5 & 20.0 & 71.8 & 28.2 \\
sociology & 201 & 83.6 & 8.0 & 82.1 & 13.9 & 81.8 & 16.4 \\
us\_foreign\_policy & 100 & 86.5 & 7.0 & 83.5 & 16.0 & 84.0 & 19.0 \\
\hline
virology & 166 & 49.7 & 12.7 & 48.8 & 23.5 & 49.7 & 22.9 \\
world\_religions & 171 & 83.9 & 7.6 & 83.3 & 12.3 & 83.0 & 11.7 \\
\hline
\end{tabular}
\caption{\label{mmlu-gemini}
Results of the sensitivity experiment across 57 MMLU subtasks for Gemini Pro, including different sensitivity settings: \textit{Token}, \textit{Order}, and \textit{Both}. \textbf{Avg. Acc} represents the mean of \(r_{forward}\) and \(r_{backward}\) accuracies for each setting.
}
\end{table*}

\begin{table*}
\centering
\small
\begin{tabular}{l|c|cr|cr|cr}
\hline
 &  & \multicolumn{2}{c|}{\textbf{Token}} & \multicolumn{2}{c|}{\textbf{Order}} & \multicolumn{2}{c}{\textbf{Both}} \\
\multicolumn{1}{c|}{\textbf{Subtask}} & \textbf{\#sample} & \textbf{Avg.} & \textbf{Fluct.} & \textbf{Avg.} & \textbf{Fluct.} & \textbf{Avg.} & \textbf{Fluct.} \\
 & & \textbf{Acc} & \textbf{Rate} & \textbf{Acc} & \textbf{Rate} & \textbf{Acc} & \textbf{Rate} \\
\hline
abstract\_algebra & 100 & 25.5 & 17.0 & 32.5 & 25.0 & 31.0 & 42.0 \\
anatomy & 135 & 63.0 & 23.7 & 62.6 & 31.9 & 64.1 & 30.4 \\
astronomy & 152 & 71.7 & 20.4 & 69.1 & 32.2 & 70.1 & 31.6 \\
business\_ethics & 100 & 57.5 & 29.0 & 64.0 & 39.0 & 62.5 & 36.0 \\
clinical\_knowledge & 265 & 67.0 & 20.8 & 67.4 & 29.1 & 69.6 & 26.4 \\
\hline
college\_biology & 144 & 69.8 & 23.6 & 72.9 & 27.1 & 71.9 & 25.7 \\
college\_chemistry & 100 & 40.5 & 32.0 & 46.5 & 34.0 & 43.0 & 43.0 \\
college\_computer\_science & 100 & 49.5 & 30.0 & 45.0 & 38.0 & 47.5 & 37.0 \\
college\_mathematics & 100 & 27.0 & 24.0 & 31.0 & 32.0 & 27.5 & 39.0 \\
college\_medicine & 173 & 56.1 & 24.9 & 55.5 & 37.6 & 57.8 & 34.7 \\
\hline
college\_physics & 102 & 29.4 & 20.6 & 35.8 & 30.4 & 34.3 & 46.1 \\
computer\_security & 100 & 70.5 & 20.0 & 69.0 & 25.0 & 74.0 & 26.0 \\
conceptual\_physics & 235 & 54.7 & 23.4 & 52.3 & 32.8 & 53.8 & 32.8 \\
econometrics & 114 & 37.7 & 29.8 & 37.3 & 43.9 & 37.3 & 43.0 \\
electrical\_engineering & 145 & 56.6 & 22.8 & 57.6 & 30.3 & 55.5 & 35.2 \\
\hline
elementary\_mathematics & 378 & 25.5 & 13.2 & 23.0 & 21.4 & 25.5 & 40.5 \\
formal\_logic & 126 & 37.3 & 22.2 & 34.5 & 42.1 & 36.5 & 55.6 \\
global\_facts & 100 & 38.0 & 50.0 & 34.5 & 37.0 & 34.0 & 57.0 \\
high\_school\_biology & 310 & 74.8 & 17.4 & 73.9 & 22.6 & 74.7 & 24.5 \\
high\_school\_chemistry & 203 & 47.5 & 25.6 & 46.8 & 32.0 & 50.5 & 36.0 \\
\hline
high\_school\_computer\_science & 100 & 60.0 & 20.0 & 62.5 & 24.0 & 64.0 & 25.0 \\
high\_school\_european\_history & 165 & 71.8 & 10.3 & 70.6 & 27.3 & 73.0 & 23.0 \\
high\_school\_geography & 198 & 79.0 & 17.2 & 77.8 & 19.2 & 78.0 & 16.2 \\
high\_school\_government\_and\_politics & 193 & 86.8 & 5.2 & 85.5 & 8.8 & 87.6 & 8.8 \\
high\_school\_macroeconomics & 390 & 58.1 & 22.8 & 56.7 & 39.2 & 57.9 & 39.0 \\
\hline
high\_school\_mathematics & 270 & 16.3 & 20.4 & 16.1 & 27.4 & 16.1 & 47.0 \\
high\_school\_microeconomics & 238 & 65.3 & 19.7 & 66.0 & 34.9 & 66.4 & 31.9 \\
high\_school\_physics & 151 & 29.8 & 23.8 & 25.2 & 34.4 & 30.1 & 46.4 \\
high\_school\_psychology & 545 & 82.1 & 13.6 & 81.6 & 21.1 & 83.6 & 15.6 \\
high\_school\_statistics & 216 & 43.3 & 27.8 & 42.4 & 39.4 & 43.8 & 47.7 \\
\hline
high\_school\_us\_history & 204 & 77.7 & 9.8 & 77.9 & 19.1 & 78.4 & 19.1 \\
high\_school\_world\_history & 237 & 78.5 & 10.1 & 77.6 & 15.6 & 79.3 & 17.7 \\
human\_aging & 223 & 67.7 & 21.1 & 66.8 & 25.1 & 67.7 & 23.8 \\
human\_sexuality & 131 & 72.9 & 18.3 & 73.3 & 25.2 & 74.0 & 23.7 \\
international\_law & 121 & 73.1 & 11.6 & 74.8 & 24.0 & 74.8 & 22.3 \\
\hline
jurisprudence & 108 & 74.1 & 12.0 & 74.1 & 19.4 & 73.1 & 13.9 \\
logical\_fallacies & 163 & 68.1 & 20.2 & 67.2 & 25.2 & 71.2 & 24.5 \\
machine\_learning & 112 & 42.4 & 17.0 & 44.6 & 37.5 & 49.1 & 31.3 \\
management & 103 & 76.2 & 9.7 & 76.2 & 24.3 & 77.7 & 19.4 \\
marketing & 234 & 85.3 & 15.8 & 86.1 & 14.9 & 88.2 & 11.5 \\
\hline
medical\_genetics & 100 & 68.5 & 21.0 & 69.0 & 25.0 & 73.0 & 24.0 \\
miscellaneous & 783 & 83.7 & 14.3 & 84.9 & 13.2 & 86.2 & 12.0 \\
moral\_disputes & 346 & 65.8 & 22.5 & 65.6 & 30.9 & 64.7 & 33.5 \\
moral\_scenarios & 895 & 21.6 & 74.7 & 21.5 & 67.6 & 22.0 & 31.2 \\
nutrition & 306 & 67.2 & 18.6 & 68.5 & 21.2 & 69.0 & 21.6 \\
\hline
philosophy & 311 & 65.9 & 21.9 & 67.0 & 27.7 & 67.8 & 28.9 \\
prehistory & 324 & 67.0 & 15.1 & 69.1 & 27.5 & 68.2 & 29.3 \\
professional\_accounting & 282 & 42.6 & 30.9 & 44.5 & 40.1 & 44.9 & 40.8 \\
professional\_law & 1534 & 45.3 & 26.4 & 45.1 & 35.0 & 46.1 & 33.5 \\
professional\_medicine & 272 & 65.3 & 20.6 & 65.4 & 29.4 & 69.7 & 23.5 \\
\hline
professional\_psychology & 612 & 63.3 & 19.4 & 63.6 & 27.6 & 64.2 & 24.7 \\
public\_relations & 110 & 64.1 & 15.5 & 61.8 & 25.5 & 65.5 & 22.7 \\
security\_studies & 245 & 64.1 & 18.8 & 66.7 & 29.4 & 63.7 & 34.7 \\
sociology & 201 & 80.3 & 14.9 & 79.4 & 17.4 & 79.1 & 20.4 \\
us\_foreign\_policy & 100 & 82.5 & 9.0 & 81.0 & 15.0 & 80.5 & 22.0 \\
\hline
virology & 166 & 49.4 & 20.5 & 50.0 & 25.3 & 48.8 & 23.5 \\
world\_religions & 171 & 79.5 & 14.0 & 80.7 & 17.5 & 83.6 & 12.3 \\
\hline
\end{tabular}
\caption{\label{mmlu-gpt-35}
Results of the sensitivity experiment across 57 MMLU subtasks for GPT-3.5, including different sensitivity settings: \textit{Token}, \textit{Order}, and \textit{Both}. \textbf{Avg. Acc} represents the mean of \(r_{forward}\) and \(r_{backward}\) accuracies for each setting.
}
\end{table*}

\begin{table*}
\centering
\small
\begin{tabular}{l|c|cr|cr|cr}
\hline
 &  & \multicolumn{2}{c|}{\textbf{Token}} & \multicolumn{2}{c|}{\textbf{Order}} & \multicolumn{2}{c}{\textbf{Both}} \\
\multicolumn{1}{c|}{\textbf{Subtask}} & \textbf{\#sample} & \textbf{Avg.} & \textbf{Fluct.} & \textbf{Avg.} & \textbf{Fluct.} & \textbf{Avg.} & \textbf{Fluct.} \\
 & & \textbf{Acc} & \textbf{Rate} & \textbf{Acc} & \textbf{Rate} & \textbf{Acc} & \textbf{Rate} \\
\hline
abstract\_algebra & 100 & 23.5 & 59.0 & 15.0 & 26.0 & 20.5 & 77.0 \\
anatomy & 135 & 36.3 & 61.5 & 35.6 & 65.9 & 37.4 & 76.3 \\
astronomy & 152 & 36.8 & 52.6 & 29.3 & 73.0 & 32.2 & 75.7 \\
business\_ethics & 100 & 33.0 & 54.0 & 31.0 & 56.0 & 33.0 & 63.0 \\
clinical\_knowledge & 265 & 38.7 & 63.8 & 36.6 & 72.5 & 37.5 & 80.8 \\
\hline
college\_biology & 144 & 37.5 & 58.3 & 31.3 & 71.5 & 33.0 & 75.0 \\
college\_chemistry & 100 & 33.5 & 68.0 & 26.0 & 61.0 & 25.5 & 81.0 \\
college\_computer\_science & 100 & 19.0 & 51.0 & 20.0 & 52.0 & 23.0 & 74.0 \\
college\_mathematics & 100 & 17.0 & 58.0 & 20.0 & 46.0 & 20.0 & 84.0 \\
college\_medicine & 173 & 33.8 & 52.0 & 29.8 & 63.0 & 31.8 & 72.3 \\
\hline
college\_physics & 102 & 17.2 & 68.6 & 16.7 & 47.1 & 15.7 & 80.4 \\
computer\_security & 100 & 33.0 & 54.0 & 35.0 & 69.0 & 39.0 & 62.0 \\
conceptual\_physics & 235 & 34.0 & 50.2 & 35.7 & 80.0 & 33.4 & 74.5 \\
econometrics & 114 & 16.2 & 44.7 & 18.4 & 53.5 & 19.7 & 62.3 \\
electrical\_engineering & 145 & 33.8 & 51.7 & 29.7 & 77.9 & 30.7 & 81.4 \\
\hline
elementary\_mathematics & 378 & 24.6 & 62.4 & 24.6 & 55.3 & 26.2 & 84.1 \\
formal\_logic & 126 & 22.6 & 68.3 & 23.0 & 42.1 & 22.6 & 89.7 \\
global\_facts & 100 & 36.5 & 57.0 & 33.5 & 56.0 & 32.0 & 67.0 \\
high\_school\_biology & 310 & 37.7 & 53.5 & 34.2 & 71.6 & 36.8 & 71.0 \\
high\_school\_chemistry & 203 & 26.8 & 65.5 & 24.9 & 66.0 & 26.1 & 83.3 \\
\hline
high\_school\_computer\_science & 100 & 30.5 & 56.0 & 33.0 & 53.0 & 34.5 & 72.0 \\
high\_school\_european\_history & 165 & 39.7 & 38.8 & 34.2 & 60.0 & 36.4 & 72.1 \\
high\_school\_geography & 198 & 38.4 & 46.5 & 35.9 & 72.2 & 36.4 & 76.3 \\
high\_school\_government\_and\_politics & 193 & 49.7 & 36.8 & 41.2 & 73.1 & 43.0 & 69.9 \\
high\_school\_macroeconomics & 390 & 35.5 & 56.2 & 30.4 & 67.2 & 32.9 & 73.6 \\
\hline
high\_school\_mathematics & 270 & 20.7 & 50.4 & 17.6 & 46.7 & 17.2 & 76.3 \\
high\_school\_microeconomics & 238 & 33.0 & 55.0 & 28.8 & 63.4 & 28.6 & 72.7 \\
high\_school\_physics & 151 & 26.2 & 65.6 & 24.8 & 53.0 & 24.2 & 83.4 \\
high\_school\_psychology & 545 & 47.2 & 48.4 & 38.6 & 70.3 & 42.0 & 71.0 \\
high\_school\_statistics & 216 & 27.5 & 61.1 & 21.5 & 53.2 & 20.8 & 75.5 \\
\hline
high\_school\_us\_history & 204 & 32.1 & 42.2 & 30.6 & 61.3 & 33.1 & 73.0 \\
high\_school\_world\_history & 237 & 37.6 & 40.1 & 39.9 & 61.2 & 39.7 & 68.4 \\
human\_aging & 223 & 32.3 & 46.6 & 38.6 & 73.1 & 38.8 & 64.1 \\
human\_sexuality & 131 & 17.9 & 26.7 & 17.6 & 33.6 & 20.6 & 33.6 \\
international\_law & 121 & 47.5 & 46.3 & 48.3 & 66.1 & 51.7 & 55.4 \\
\hline
jurisprudence & 108 & 36.6 & 51.9 & 40.7 & 67.6 & 38.4 & 77.8 \\
logical\_fallacies & 163 & 41.4 & 53.4 & 39.6 & 76.1 & 39.9 & 77.3 \\
machine\_learning & 112 & 23.7 & 44.6 & 31.3 & 74.1 & 30.8 & 84.8 \\
management & 103 & 41.7 & 57.3 & 33.5 & 82.5 & 35.9 & 72.8 \\
marketing & 234 & 41.7 & 53.0 & 44.0 & 86.3 & 49.1 & 70.5 \\
\hline
medical\_genetics & 100 & 30.5 & 60.0 & 31.0 & 84.0 & 32.5 & 76.0 \\
miscellaneous & 783 & 43.2 & 51.9 & 43.8 & 68.5 & 45.9 & 67.0 \\
moral\_disputes & 346 & 29.3 & 44.2 & 28.5 & 73.7 & 30.5 & 69.7 \\
moral\_scenarios & 895 & 6.0 & 15.8 & 6.0 & 2.0 & 5.9 & 31.3 \\
nutrition & 306 & 35.5 & 62.4 & 33.7 & 69.3 & 39.2 & 76.5 \\
\hline
philosophy & 311 & 40.0 & 49.5 & 36.7 & 76.8 & 40.0 & 67.5 \\
prehistory & 324 & 37.3 & 53.4 & 36.1 & 77.5 & 37.0 & 75.3 \\
professional\_accounting & 282 & 26.2 & 56.4 & 27.7 & 64.9 & 27.5 & 72.0 \\
professional\_law & 1534 & 28.3 & 30.9 & 26.7 & 53.1 & 25.1 & 59.6 \\
professional\_medicine & 272 & 32.4 & 51.8 & 25.2 & 51.5 & 21.7 & 79.4 \\
\hline
professional\_psychology & 612 & 33.2 & 50.7 & 32.7 & 71.7 & 34.2 & 67.0 \\
public\_relations & 110 & 43.6 & 40.0 & 39.5 & 80.0 & 41.4 & 73.6 \\
security\_studies & 245 & 40.0 & 36.7 & 32.4 & 82.0 & 32.7 & 81.2 \\
sociology & 201 & 46.0 & 39.3 & 40.8 & 73.1 & 44.8 & 69.7 \\
us\_foreign\_policy & 100 & 42.0 & 47.0 & 40.0 & 66.0 & 39.0 & 63.0 \\
\hline
virology & 166 & 34.3 & 51.2 & 31.6 & 77.7 & 30.7 & 65.1 \\
world\_religions & 171 & 33.3 & 55.6 & 39.8 & 74.9 & 44.4 & 67.3 \\
\hline
\end{tabular}
\caption{\label{mmlu-llama-7b}
Results of the sensitivity experiment across 57 MMLU subtasks for  LLaMA 2 7B, including different sensitivity settings: \textit{Token}, \textit{Order}, and \textit{Both}. \textbf{Avg. Acc} represents the mean of \(r_{forward}\) and \(r_{backward}\) accuracies for each setting.
}
\end{table*}

\begin{table*}
\centering
\small
\begin{tabular}{l|c|cr|cr|cr}
\hline
 &  & \multicolumn{2}{c|}{\textbf{Token}} & \multicolumn{2}{c|}{\textbf{Order}} & \multicolumn{2}{c}{\textbf{Both}} \\
\multicolumn{1}{c|}{\textbf{Subtask}} & \textbf{\#sample} & \textbf{Avg.} & \textbf{Fluct.} & \textbf{Avg.} & \textbf{Fluct.} & \textbf{Avg.} & \textbf{Fluct.} \\
 & & \textbf{Acc} & \textbf{Rate} & \textbf{Acc} & \textbf{Rate} & \textbf{Acc} & \textbf{Rate} \\
\hline
abstract\_algebra & 100 & 26.5 & 7.0 & 25.5 & 35.0 & 26.0 & 29.0 \\
anatomy & 135 & 41.5 & 43.7 & 41.1 & 34.1 & 42.6 & 48.9 \\
astronomy & 152 & 38.5 & 38.2 & 45.1 & 42.8 & 39.8 & 50.7 \\
business\_ethics & 100 & 37.0 & 47.0 & 38.0 & 36.0 & 36.0 & 55.0 \\
clinical\_knowledge & 265 & 40.0 & 48.7 & 42.5 & 38.1 & 40.0 & 50.2 \\
\hline
college\_biology & 144 & 40.3 & 45.1 & 39.9 & 38.2 & 41.3 & 45.1 \\
college\_chemistry & 100 & 31.5 & 52.0 & 22.5 & 41.0 & 23.5 & 60.0 \\
college\_computer\_science & 100 & 30.0 & 40.0 & 29.5 & 47.0 & 29.5 & 62.0 \\
college\_mathematics & 100 & 24.0 & 33.0 & 19.0 & 38.0 & 22.0 & 70.0 \\
college\_medicine & 173 & 31.8 & 41.6 & 32.4 & 35.8 & 29.5 & 53.8 \\
\hline
college\_physics & 102 & 16.7 & 34.3 & 21.6 & 43.1 & 23.0 & 57.8 \\
computer\_security & 100 & 44.5 & 30.0 & 47.5 & 41.0 & 48.5 & 47.0 \\
conceptual\_physics & 235 & 32.1 & 31.9 & 31.5 & 36.2 & 33.8 & 42.1 \\
econometrics & 114 & 28.5 & 46.5 & 29.8 & 49.1 & 28.1 & 56.1 \\
electrical\_engineering & 145 & 31.0 & 37.9 & 30.3 & 47.6 & 31.0 & 57.2 \\
\hline
elementary\_mathematics & 378 & 24.9 & 37.0 & 24.5 & 34.7 & 25.0 & 59.3 \\
formal\_logic & 126 & 22.2 & 40.5 & 23.0 & 51.6 & 21.4 & 84.9 \\
global\_facts & 100 & 31.5 & 54.0 & 29.5 & 24.0 & 32.0 & 63.0 \\
high\_school\_biology & 310 & 45.3 & 33.9 & 44.5 & 42.3 & 43.5 & 51.0 \\
high\_school\_chemistry & 203 & 30.8 & 41.4 & 28.6 & 43.8 & 30.3 & 53.2 \\
\hline
high\_school\_computer\_science & 100 & 39.0 & 20.0 & 40.5 & 37.0 & 37.5 & 47.0 \\
high\_school\_european\_history & 165 & 47.9 & 45.5 & 47.6 & 43.6 & 44.8 & 52.7 \\
high\_school\_geography & 198 & 44.4 & 37.4 & 45.5 & 39.4 & 41.7 & 55.6 \\
high\_school\_government\_and\_politics & 193 & 54.9 & 31.6 & 56.2 & 32.1 & 52.8 & 43.5 \\
high\_school\_macroeconomics & 390 & 35.6 & 37.9 & 35.6 & 40.8 & 34.0 & 47.7 \\
\hline
high\_school\_mathematics & 270 & 18.1 & 30.7 & 18.5 & 25.2 & 18.7 & 54.1 \\
high\_school\_microeconomics & 238 & 40.5 & 34.9 & 33.8 & 46.2 & 37.0 & 47.9 \\
high\_school\_physics & 151 & 30.1 & 31.8 & 27.8 & 41.7 & 26.8 & 58.3 \\
high\_school\_psychology & 545 & 47.2 & 35.4 & 46.4 & 34.3 & 45.3 & 46.8 \\
high\_school\_statistics & 216 & 27.3 & 34.3 & 26.2 & 44.4 & 26.2 & 59.3 \\
\hline
high\_school\_us\_history & 204 & 54.2 & 38.2 & 53.2 & 41.7 & 51.5 & 51.0 \\
high\_school\_world\_history & 237 & 50.4 & 35.4 & 50.8 & 40.9 & 52.3 & 48.5 \\
human\_aging & 223 & 40.6 & 39.5 & 35.4 & 37.2 & 41.3 & 48.4 \\
human\_sexuality & 131 & 21.4 & 19.8 & 22.9 & 19.1 & 23.7 & 32.8 \\
international\_law & 121 & 63.2 & 29.8 & 63.6 & 35.5 & 64.0 & 39.7 \\
\hline
jurisprudence & 108 & 49.5 & 38.0 & 46.8 & 39.8 & 47.7 & 50.9 \\
logical\_fallacies & 163 & 44.2 & 37.4 & 48.2 & 44.8 & 43.6 & 47.2 \\
machine\_learning & 112 & 28.6 & 26.8 & 27.2 & 37.5 & 30.4 & 38.4 \\
management & 103 & 42.7 & 38.8 & 40.8 & 39.8 & 41.7 & 46.6 \\
marketing & 234 & 53.4 & 45.3 & 55.6 & 36.8 & 55.3 & 47.9 \\
\hline
medical\_genetics & 100 & 35.0 & 39.0 & 33.5 & 42.0 & 34.5 & 52.0 \\
miscellaneous & 783 & 49.7 & 36.0 & 48.9 & 31.2 & 52.8 & 51.5 \\
moral\_disputes & 346 & 37.7 & 37.0 & 37.0 & 41.9 & 38.0 & 47.1 \\
moral\_scenarios & 895 & 18.4 & 23.6 & 18.2 & 6.8 & 18.6 & 80.1 \\
nutrition & 306 & 38.1 & 45.8 & 37.9 & 38.9 & 40.4 & 52.3 \\
\hline
philosophy & 311 & 45.3 & 23.5 & 42.4 & 39.9 & 43.2 & 41.5 \\
prehistory & 324 & 42.4 & 43.2 & 42.7 & 34.0 & 42.4 & 49.1 \\
professional\_accounting & 282 & 31.7 & 47.2 & 31.7 & 41.8 & 31.6 & 54.6 \\
professional\_law & 1534 & 30.7 & 34.5 & 30.9 & 43.9 & 30.1 & 57.0 \\
professional\_medicine & 272 & 31.4 & 39.3 & 30.9 & 40.8 & 30.0 & 57.4 \\
\hline
professional\_psychology & 612 & 38.4 & 41.3 & 39.1 & 37.6 & 38.4 & 50.3 \\
public\_relations & 110 & 44.5 & 35.5 & 44.5 & 34.5 & 42.3 & 47.3 \\
security\_studies & 245 & 38.0 & 39.6 & 38.2 & 43.3 & 37.6 & 51.0 \\
sociology & 201 & 49.0 & 34.8 & 50.0 & 35.8 & 49.0 & 40.3 \\
us\_foreign\_policy & 100 & 55.5 & 38.0 & 48.5 & 42.0 & 49.5 & 41.0 \\
\hline
virology & 166 & 37.0 & 32.5 & 32.5 & 37.3 & 34.9 & 41.6 \\
world\_religions & 171 & 43.0 & 37.4 & 50.0 & 32.7 & 53.2 & 43.3 \\
\hline
\end{tabular}
\caption{\label{mmlu-llama-13b}
Results of the sensitivity experiment across 57 MMLU subtasks for  LLaMA 2 13B, including different sensitivity settings: \textit{Token}, \textit{Order}, and \textit{Both}. \textbf{Avg. Acc} represents the mean of \(r_{forward}\) and \(r_{backward}\) accuracies for each setting.
}
\end{table*}

\begin{table*}
\centering
\small
\begin{tabular}{l|c|cr|cr|cr}
\hline
 &  & \multicolumn{2}{c|}{\textbf{Token}} & \multicolumn{2}{c|}{\textbf{Order}} & \multicolumn{2}{c}{\textbf{Both}} \\
\multicolumn{1}{c|}{\textbf{Subtask}} & \textbf{\#sample} & \textbf{Avg.} & \textbf{Fluct.} & \textbf{Avg.} & \textbf{Fluct.} & \textbf{Avg.} & \textbf{Fluct.} \\
 & & \textbf{Acc} & \textbf{Rate} & \textbf{Acc} & \textbf{Rate} & \textbf{Acc} & \textbf{Rate} \\
\hline
abstract\_algebra & 100 & 36.0 & 41.0 & 36.5 & 71.0 & 33.5 & 58.0 \\
anatomy & 135 & 41.9 & 57.0 & 44.1 & 65.2 & 46.7 & 39.3 \\
astronomy & 152 & 44.7 & 60.5 & 53.9 & 50.7 & 55.9 & 40.8 \\
business\_ethics & 100 & 45.5 & 60.0 & 44.5 & 48.0 & 46.5 & 57.0 \\
clinical\_knowledge & 265 & 49.4 & 57.0 & 53.8 & 49.4 & 54.0 & 40.0 \\
\hline
college\_biology & 144 & 50.7 & 55.6 & 49.7 & 53.5 & 51.7 & 41.0 \\
college\_chemistry & 100 & 30.5 & 76.0 & 40.5 & 76.0 & 29.5 & 53.0 \\
college\_computer\_science & 100 & 38.5 & 64.0 & 41.0 & 57.0 & 38.0 & 53.0 \\
college\_mathematics & 100 & 31.0 & 63.0 & 32.0 & 68.0 & 26.5 & 61.0 \\
college\_medicine & 173 & 39.3 & 50.9 & 42.2 & 46.8 & 40.2 & 50.3 \\
\hline
college\_physics & 102 & 25.5 & 67.6 & 32.4 & 76.5 & 24.5 & 55.9 \\
computer\_security & 100 & 58.5 & 36.0 & 53.0 & 48.0 & 55.0 & 35.0 \\
conceptual\_physics & 235 & 40.4 & 46.4 & 40.4 & 50.2 & 39.8 & 45.1 \\
econometrics & 114 & 30.3 & 77.2 & 28.9 & 51.8 & 32.9 & 67.5 \\
electrical\_engineering & 145 & 39.3 & 62.8 & 43.8 & 57.2 & 40.3 & 48.3 \\
\hline
elementary\_mathematics & 378 & 31.7 & 72.0 & 32.8 & 77.0 & 32.4 & 55.0 \\
formal\_logic & 126 & 23.8 & 53.2 & 28.2 & 65.9 & 22.6 & 50.8 \\
global\_facts & 100 & 30.0 & 77.0 & 26.0 & 86.0 & 30.0 & 59.0 \\
high\_school\_biology & 310 & 54.7 & 49.0 & 58.4 & 41.0 & 56.9 & 44.2 \\
high\_school\_chemistry & 203 & 31.0 & 77.8 & 36.7 & 71.9 & 36.5 & 46.3 \\
\hline
high\_school\_computer\_science & 100 & 50.0 & 58.0 & 45.0 & 52.0 & 52.0 & 49.0 \\
high\_school\_european\_history & 165 & 60.0 & 40.6 & 58.5 & 25.5 & 55.2 & 49.1 \\
high\_school\_geography & 198 & 56.1 & 53.5 & 62.6 & 36.4 & 60.9 & 39.9 \\
high\_school\_government\_and\_politics & 193 & 64.5 & 38.3 & 66.6 & 35.2 & 68.9 & 26.9 \\
high\_school\_macroeconomics & 390 & 45.0 & 57.2 & 49.9 & 57.2 & 49.1 & 40.5 \\
\hline
high\_school\_mathematics & 270 & 27.0 & 76.3 & 31.3 & 71.9 & 28.3 & 65.9 \\
high\_school\_microeconomics & 238 & 45.0 & 62.6 & 50.8 & 60.1 & 50.4 & 40.3 \\
high\_school\_physics & 151 & 30.8 & 66.9 & 33.4 & 69.5 & 29.8 & 52.3 \\
high\_school\_psychology & 545 & 62.1 & 45.3 & 66.5 & 32.5 & 66.1 & 35.2 \\
high\_school\_statistics & 216 & 30.8 & 65.7 & 42.6 & 64.4 & 34.0 & 41.2 \\
\hline
high\_school\_us\_history & 204 & 58.8 & 47.5 & 62.7 & 31.9 & 61.3 & 45.6 \\
high\_school\_world\_history & 237 & 62.7 & 41.4 & 62.0 & 32.9 & 64.1 & 38.0 \\
human\_aging & 223 & 50.9 & 59.2 & 41.9 & 53.8 & 50.2 & 45.3 \\
human\_sexuality & 131 & 33.6 & 30.5 & 33.6 & 19.8 & 34.7 & 32.8 \\
international\_law & 121 & 64.5 & 46.3 & 62.8 & 44.6 & 68.2 & 33.1 \\
\hline
jurisprudence & 108 & 56.0 & 47.2 & 56.9 & 40.7 & 60.2 & 36.1 \\
logical\_fallacies & 163 & 54.0 & 49.7 & 54.6 & 39.9 & 58.9 & 39.3 \\
machine\_learning & 112 & 34.4 & 43.8 & 30.4 & 44.6 & 36.2 & 30.4 \\
management & 103 & 52.4 & 49.5 & 61.7 & 33.0 & 58.3 & 44.7 \\
marketing & 234 & 68.4 & 43.6 & 64.3 & 31.6 & 71.8 & 35.0 \\
\hline
medical\_genetics & 100 & 49.0 & 62.0 & 47.0 & 56.0 & 49.5 & 45.0 \\
miscellaneous & 783 & 67.6 & 37.0 & 66.5 & 35.1 & 70.6 & 31.5 \\
moral\_disputes & 346 & 48.1 & 63.3 & 47.8 & 52.6 & 51.9 & 34.4 \\
moral\_scenarios & 895 & 26.1 & 34.1 & 23.9 & 24.5 & 22.5 & 46.9 \\
nutrition & 306 & 46.4 & 58.2 & 47.7 & 51.6 & 49.7 & 39.9 \\
\hline
philosophy & 311 & 51.9 & 59.2 & 55.0 & 49.8 & 58.8 & 34.7 \\
prehistory & 324 & 51.7 & 58.0 & 52.5 & 51.9 & 59.6 & 37.0 \\
professional\_accounting & 282 & 32.3 & 71.6 & 32.6 & 69.5 & 37.1 & 46.5 \\
professional\_law & 1534 & 33.6 & 53.8 & 34.0 & 50.3 & 36.0 & 47.3 \\
professional\_medicine & 272 & 28.3 & 54.4 & 38.6 & 41.9 & 33.5 & 61.4 \\
\hline
professional\_psychology & 612 & 47.0 & 56.7 & 44.1 & 53.1 & 50.4 & 39.1 \\
public\_relations & 110 & 50.9 & 47.3 & 48.2 & 48.2 & 50.9 & 44.5 \\
security\_studies & 245 & 43.7 & 72.7 & 50.4 & 50.6 & 47.8 & 44.1 \\
sociology & 201 & 61.4 & 48.8 & 63.2 & 40.8 & 65.7 & 31.8 \\
us\_foreign\_policy & 100 & 69.0 & 36.0 & 68.0 & 38.0 & 70.0 & 36.0 \\
\hline
virology & 166 & 38.9 & 60.2 & 40.4 & 39.2 & 39.5 & 40.4 \\
world\_religions & 171 & 66.7 & 37.4 & 64.9 & 33.9 & 69.0 & 23.4 \\
\hline
\end{tabular}
\caption{\label{mmlu-llama-70b}
Results of the sensitivity experiment across 57 MMLU subtasks for  LLaMA 2 70B, including different sensitivity settings: \textit{Token}, \textit{Order}, and \textit{Both}. \textbf{Avg. Acc} represents the mean of \(r_{forward}\) and \(r_{backward}\) accuracies for each setting.
}
\end{table*}

\begin{table*}[h]
\centering
\small
\begin{tabular}{l|c|c|c|c}
\hline
 & A (\%) & B (\%) & C (\%) & D (\%) \\
\hline
Ground truth & 25.04 & 24.75 & \textbf{25.73} & 24.48 \\
\hline
PaLM 2 & 23.47 & 24.54 & \textbf{26.26} & 25.73 \\
Gemini Pro & 15.94 & 27.96 & \textbf{30.17} & 25.92 \\
GPT 3.5 & 18.98 & \textbf{32.70} & 27.13 & 21.19 \\
LLaMA2-7B & 0.05 & \textbf{55.56} & 27.16 & 17.23 \\
LLaMA2-13B & 1.17 & \textbf{65.41} & 28.82 & 4.60 \\
LLaMA2-70B & 11.36 & \textbf{38.78} & 33.93 & 15.93 \\
\hline
\end{tabular}
\caption{\label{option-portion-first}Option proportion statistics and ground truth
label proportions for the HellaSwag dataset. The most frequent option in each row is highlighted in \textbf{bold}.}
\end{table*}

\begin{table*}[h]
\centering
\small
\begin{tabular}{l|c|c|c|c}
\hline
 & A (\%) & B (\%) & C (\%) & D (\%) \\
\hline
Ground truth & 22.95 & 24.65 & 25.51 & \textbf{26.88} \\
\hline
PaLM 2 & 15.32 & \textbf{29.30} & 29.17 & 26.20 \\
Gemini Pro & 13.74 & 26.85 & 28.79 & \textbf{30.62} \\
GPT 3.5 & 21.32 & \textbf{30.55} & 27.51 & 20.62 \\
LLaMA2-7B & \textbf{47.82} & 26.34 & 22.65 & 3.20 \\
LLaMA2-13B & 10.66 & \textbf{43.65} & 34.85 & 10.84 \\
LLaMA2-70B & 11.30 & 35.93 & \textbf{36.60} & 16.17 \\
\hline
\end{tabular}
\caption{Option proportion statistics and ground truth
label proportions for the MMLU dataset. The most frequent option in each row is highlighted in \textbf{bold}.}
\end{table*}

\begin{table*}[h]
\centering
\small
\begin{tabular}{l|c|c|c|c}
\hline
 & A (\%) & B (\%) & C (\%) & D (\%) \\
\hline
Ground truth & \textbf{27.60} & 25.20 & 26.40 & 20.80 \\
\hline
PaLM 2 & 24.30 & \textbf{26.71} & 26.31 & 22.69 \\
Gemini Pro & 22.60 & 26.00 & \textbf{29.80} & 21.60 \\
GPT 3.5 & 19.88 & 31.73 & \textbf{29.92} & 18.47 \\
LLaMA2-7B & \textbf{69.15} & 14.11 & 16.73 & 0.00 \\
LLaMA2-13B & 1.04 & 35.07 & \textbf{53.65} & 10.23 \\
LLaMA2-70B & 8.01 & 31.21 & \textbf{39.84} & 20.94 \\
\hline
\end{tabular}
\caption{Option proportion statistics and ground truth
label proportions for the OpenBookQA dataset. The most frequent option in each row is highlighted in \textbf{bold}.}
\end{table*}

\begin{table*}[h]
\centering
\small
\begin{tabular}{l|c|c}
\hline
 & A (\%) & B (\%) \\
\hline
Ground truth & 49.57 & \textbf{50.43} \\
\hline
PaLM 2 & 30.59 & \textbf{69.41} \\
Gemini Pro & 19.02 & \textbf{80.98} \\
GPT 3.5 & 41.48 & \textbf{58.52} \\
LLaMA2-7B & 0.08 & \textbf{99.92} \\
LLaMA2-13B & 0.53 & \textbf{99.47} \\
LLaMA2-70B & \textbf{100.00} & 0.00 \\
\hline
\end{tabular}
\caption{Option proportion statistics and ground truth
label proportions for the Winogrande dataset. The most frequent option in each row is highlighted in \textbf{bold}.}
\end{table*}

\begin{table*}[h]
\centering
\small
\begin{tabular}{l|c|c|c|c|c}
\hline
 & A (\%) & B (\%) & C (\%) & D (\%)  & E (\%) \\
\hline
Ground truth & 20.80 & 20.27 & \textbf{22.58} & 20.94 & 15.41 \\
\hline
PaLM 2 & 0.67 & 18.50 & \textbf{37.56} & 27.19 & 16.08 \\
Gemini Pro & 1.21 & 20.74 & \textbf{30.59} & 24.46 & 23.02 \\
GPT 3.5 & 1.99 & 22.90 & \textbf{45.73} & 25.33 & 4.05 \\
LLaMA2-7B & \textbf{38.21} & 24.51 & 32.32 & 4.96 & 0.00 \\
LLaMA2-13B & 0.27 & \textbf{57.29} & 37.61 & 3.65 & 1.18 \\
LLaMA2-70B & 0.24 & 14.10 & \textbf{74.52} & 7.05 & 4.08 \\
\hline
\end{tabular}
\caption{\label{option-portion-last}Option proportion statistics and ground truth
label proportions for the MathQA dataset. The most frequent option in each row is highlighted in \textbf{bold}.}
\end{table*}

\begin{table*}
\centering
\definecolor{darkgreen}{rgb}{0.0, 0.5, 0.0}
\small
\begin{tabular}{l|c|cr|cr|cr}
\hline
 &  & \multicolumn{2}{c|}{\textbf{Token}} & \multicolumn{2}{c|}{\textbf{Order}} & \multicolumn{2}{c}{\textbf{Both}} \\
\multicolumn{1}{c|}{\textbf{Subtask}} & \textbf{Subcategory} & \textbf{Acc} & \textbf{Diff} & \textbf{Acc} & \textbf{Diff} & \textbf{Acc} & \textbf{Diff} \\
\hline
abstract\_algebra& STEM & 27.00 & 1.50 & 37.00 & 4.50 & 39.00 & 8.00 \\
us\_foreign\_policy& Social Sciences & 83.00 & 0.50 & 84.00 & 3.00 & 88.00 & 7.50 \\
college\_chemistry& STEM & 43.00 & 2.50 & 47.00 & 0.50 & 50.00 & 7.00 \\
elementary\_mathematics& STEM & 29.89 & 4.37 & 30.69 & 7.67 & 32.01 & 6.48 \\
computer\_security& STEM & 73.00 & 2.50 & 76.00 & 7.00 & 80.00 & 6.00 \\
high\_school\_computer\_science& STEM & 66.00 & 6.00 & 65.00 & 2.50 & 70.00 & 6.00 \\
global\_facts& Other & 41.00 & 3.00 & 35.00 & 0.50 & 39.00 & 5.00 \\
moral\_disputes& Humanities & 68.79 & 3.03 & 68.79 & 3.18 & 69.36 & 4.62 \\
astronomy& STEM & 75.00 & 3.29 & 75.00 & 5.92 & 74.34 & 4.28 \\
high\_school\_european\_history& Humanities & 72.84 & 1.02 & 76.54 & 5.94 & 77.16 & 4.13 \\
professional\_psychology& Social Sciences & 65.85 & 2.53 & 66.01 & 2.45 & 68.30 & 4.08 \\
prehistory& Humanities & 68.52 & 1.54 & 74.07 & 4.94 & 72.22 & 4.01 \\
medical\_genetics& Other & 72.00 & 3.50 & 75.00 & 6.00 & 77.00 & 4.00 \\
formal\_logic& Humanities & 38.10 & 0.79 & 37.30 & 2.78 & 40.48 & 3.97 \\
econometrics& Social Sciences & 38.60 & 0.88 & 40.35 & 3.07 & 41.23 & 3.95 \\
management& Other & 78.64 & 2.43 & 78.64 & 2.43 & 81.55 & 3.88 \\
high\_school\_biology& STEM & 77.42 & 2.58 & 77.42 & 3.55 & 78.39 & 3.71 \\
jurisprudence& Humanities & 75.93 & 1.85 & 76.85 & 2.78 & 76.85 & 3.70 \\
security\_studies& Social Sciences & 65.71 & 1.63 & 66.53 & -0.20 & 67.35 & 3.67 \\
philosophy& Humanities & 71.38 & 5.47 & 71.38 & 4.34 & 71.38 & 3.54 \\
high\_school\_us\_history& Humanities & 79.90 & 2.21 & 81.86 & 3.92 & 81.86 & 3.43 \\
nutrition& Other & 69.28 & 2.12 & 72.22 & 3.76 & 72.22 & 3.27 \\
high\_school\_chemistry& STEM & 50.25 & 2.71 & 50.25 & 3.45 & 53.69 & 3.20 \\
conceptual\_physics& STEM & 57.87 & 3.19 & 55.32 & 2.98 & 57.02 & 3.19 \\
high\_school\_statistics& STEM & 45.83 & 2.55 & 47.22 & 4.86 & 46.76 & 3.01 \\
high\_school\_world\_history& Humanities & 79.32 & 0.84 & 79.75 & 2.11 & 82.28 & 2.95 \\
college\_physics& STEM & 34.31 & 4.90 & 38.24 & 2.45 & 37.25 & 2.94 \\
human\_aging& Other & 67.71 & 0.00 & 68.61 & 1.79 & 70.40 & 2.69 \\
high\_school\_mathematics& STEM & 18.52 & 2.22 & 21.11 & 5.00 & 18.52 & 2.41 \\
high\_school\_macroeconomics& Social Sciences & 58.46 & 0.38 & 60.00 & 3.33 & 60.26 & 2.31 \\
high\_school\_microeconomics& Social Sciences & 68.07 & 2.73 & 67.23 & 1.26 & 68.49 & 2.10 \\
international\_law& Humanities & 73.55 & 0.41 & 76.03 & 1.24 & 76.86 & 2.07 \\
moral\_scenarios& Humanities & 23.24 & 1.62 & 17.99 & -3.46 & 23.91 & 1.96 \\
professional\_law& Humanities & 46.94 & 1.66 & 47.00 & 1.92 & 47.98 & 1.86 \\
public\_relations& Social Sciences & 65.45 & 1.36 & 65.45 & 3.64 & 67.27 & 1.82 \\
miscellaneous& Other & 86.59 & 2.87 & 87.10 & 2.17 & 87.99 & 1.79 \\
high\_school\_geography& Social Sciences & 81.31 & 2.27 & 80.30 & 2.53 & 79.80 & 1.77 \\
college\_medicine& Other & 58.38 & 2.31 & 58.96 & 3.47 & 59.54 & 1.73 \\
high\_school\_physics& STEM & 34.44 & 4.64 & 29.80 & 4.64 & 31.79 & 1.66 \\
human\_sexuality& Social Sciences & 76.34 & 3.44 & 75.57 & 2.29 & 75.57 & 1.53 \\
college\_computer\_science& STEM & 55.00 & 5.50 & 51.00 & 6.00 & 49.00 & 1.50 \\
marketing& Other & 88.03 & 2.78 & 88.46 & 2.35 & 89.74 & 1.50 \\
professional\_medicine& Other & 68.01 & 2.76 & 70.59 & 5.15 & 70.96 & 1.29 \\
anatomy& STEM & 66.67 & 3.70 & 67.41 & 4.81 & 65.19 & 1.11 \\
college\_biology& STEM & 72.22 & 2.43 & 72.92 & -0.00 & 72.92 & 1.04 \\
electrical\_engineering& STEM & 58.62 & 2.07 & 57.24 & -0.34 & 56.55 & 1.03 \\
sociology& Social Sciences & 81.59 & 1.24 & 82.59 & 3.23 & 80.10 & 1.00 \\
clinical\_knowledge& Other & 68.30 & 1.32 & 69.43 & 2.08 & 70.57 & 0.94 \\
high\_school\_psychology& Social Sciences & 84.40 & 2.29 & 85.87 & 4.31 & 84.22 & 0.64 \\
professional\_accounting& Other & 44.68 & 2.13 & 47.16 & 2.66 & 45.39 & 0.53 \\
high\_school\_government\_and\_politics& Social Sciences & 87.56 & 0.78 & 87.05 & 1.55 & 88.08 & 0.52 \\
business\_ethics& Other & 61.00 & 3.50 & 61.00 & -3.00 & 63.00 & 0.50 \\
college\_mathematics& STEM & 27.00 & 0.00 & 35.00 & 4.00 & 28.00 & 0.50 \\
logical\_fallacies& Humanities & 69.94 & 1.84 & 68.10 & 0.92 & 71.17 & 0.00 \\
world\_religions& Humanities & 82.46 & 2.92 & 83.63 & 2.92 & 82.46 & -1.17 \\
virology& Other & 48.80 & -0.60 & 48.80 & -1.20 & 46.39 & -2.41 \\
machine\_learning& STEM & 44.64 & 2.23 & 45.54 & 0.89 & 45.54 & -3.57 \\
\hline
\end{tabular}
\caption{\label{mmlu-gray-box-weighting}
Results of probability weighting method of GPT-3.5 model across 57 MMLU subtask.
}
\end{table*}

\begin{table*}
\centering
\small
\begin{tabular}{l|c|cr|cr|cr}
\hline
 &  & \multicolumn{2}{c|}{\textbf{Token}} & \multicolumn{2}{c|}{\textbf{Order}} & \multicolumn{2}{c}{\textbf{Both}} \\
\multicolumn{1}{c|}{\textbf{Subtask}} & \textbf{Subcategory} & \textbf{Acc} & \textbf{Diff} & \textbf{Acc} & \textbf{Diff} & \textbf{Acc} & \textbf{Diff} \\
\hline
elementary\_mathematics & STEM & 39.42 & 13.89 & 36.51 & 13.49 & 39.81 & 14.29 \\
high\_school\_mathematics & STEM & 28.89 & 12.59 & 26.67 & 10.56 & 28.33 & 12.22 \\
college\_physics & STEM & 41.18 & 11.76 & 43.63 & 7.84 & 45.59 & 11.27 \\
college\_chemistry & STEM & 44.00 & 3.50 & 49.00 & 2.50 & 50.00 & 7.00 \\
formal\_logic & Humanities & 41.67 & 4.37 & 38.49 & 3.97 & 40.08 & 3.57 \\
college\_computer\_science & STEM & 51.50 & 2.00 & 47.00 & 2.00 & 51.00 & 3.50 \\
high\_school\_statistics & STEM & 47.92 & 4.63 & 44.68 & 2.31 & 46.99 & 3.24 \\
college\_mathematics & STEM & 28.50 & 1.50 & 31.50 & 0.50 & 29.50 & 2.00 \\
global\_facts & Other & 40.50 & 2.50 & 34.00 & -0.50 & 36.00 & 2.00 \\
abstract\_algebra & STEM & 29.00 & 3.50 & 33.00 & 0.50 & 33.00 & 2.00 \\
high\_school\_physics & STEM & 31.46 & 1.66 & 29.14 & 3.97 & 32.12 & 1.99 \\
econometrics & Social Sciences & 37.72 & 0.00 & 37.72 & 0.44 & 39.04 & 1.75 \\
high\_school\_chemistry & STEM & 50.00 & 2.46 & 49.26 & 2.46 & 52.22 & 1.72 \\
computer\_security & STEM & 71.00 & 0.50 & 70.50 & 1.50 & 75.50 & 1.50 \\
professional\_accounting & Other & 44.50 & 1.95 & 45.39 & 0.89 & 46.28 & 1.42 \\
college\_medicine & Other & 56.94 & 0.87 & 56.65 & 1.16 & 58.96 & 1.16 \\
anatomy & STEM & 64.07 & 1.11 & 64.07 & 1.48 & 65.19 & 1.11 \\
high\_school\_microeconomics & Social Sciences & 65.55 & 0.21 & 66.81 & 0.84 & 67.44 & 1.05 \\
medical\_genetics & Other & 70.50 & 2.00 & 72.00 & 3.00 & 74.00 & 1.00 \\
astronomy & STEM & 72.04 & 0.33 & 70.07 & 0.99 & 71.05 & 0.99 \\
nutrition & Other & 67.81 & 0.65 & 70.10 & 1.63 & 69.93 & 0.98 \\
high\_school\_biology & STEM & 75.48 & 0.65 & 75.00 & 1.13 & 75.65 & 0.97 \\
security\_studies & Social Sciences & 64.49 & 0.41 & 67.55 & 0.82 & 64.49 & 0.82 \\
clinical\_knowledge & Other & 68.11 & 1.13 & 68.30 & 0.94 & 70.38 & 0.75 \\
high\_school\_us\_history & Humanities & 77.70 & 0.00 & 78.43 & 0.49 & 79.17 & 0.74 \\
high\_school\_psychology & Social Sciences & 82.39 & 0.28 & 82.11 & 0.55 & 84.31 & 0.73 \\
philosophy & Humanities & 66.24 & 0.32 & 67.04 & 0.00 & 68.49 & 0.64 \\
high\_school\_macroeconomics & Social Sciences & 59.36 & 1.28 & 57.31 & 0.64 & 58.59 & 0.64 \\
conceptual\_physics & STEM & 55.11 & 0.43 & 52.77 & 0.43 & 54.47 & 0.64 \\
professional\_medicine & Other & 65.62 & 0.37 & 65.99 & 0.55 & 70.22 & 0.55 \\
high\_school\_geography & Social Sciences & 79.55 & 0.51 & 77.78 & 0.00 & 78.54 & 0.51 \\
high\_school\_computer\_science & STEM & 61.50 & 1.50 & 63.50 & 1.00 & 64.50 & 0.50 \\
professional\_law & Humanities & 45.31 & 0.03 & 45.24 & 0.16 & 46.61 & 0.49 \\
prehistory & Humanities & 66.67 & -0.31 & 68.21 & -0.93 & 68.67 & 0.46 \\
high\_school\_european\_history & Humanities & 72.22 & 0.40 & 71.60 & 1.00 & 73.46 & 0.43 \\
college\_biology & STEM & 70.49 & 0.69 & 72.57 & -0.35 & 72.22 & 0.35 \\
virology & Other & 49.10 & -0.30 & 49.70 & -0.30 & 49.10 & 0.30 \\
world\_religions & Humanities & 79.24 & -0.29 & 81.58 & 0.88 & 83.92 & 0.29 \\
marketing & Other & 85.26 & 0.00 & 86.11 & 0.00 & 88.46 & 0.21 \\
miscellaneous & Other & 84.29 & 0.57 & 85.19 & 0.26 & 86.40 & 0.19 \\
moral\_scenarios & Humanities & 21.62 & 0.00 & 21.34 & -0.11 & 22.12 & 0.17 \\
international\_law & Humanities & 73.55 & 0.41 & 75.21 & 0.41 & 74.79 & 0.00 \\
human\_sexuality & Social Sciences & 73.28 & 0.38 & 72.52 & -0.76 & 74.05 & 0.00 \\
logical\_fallacies & Humanities & 67.79 & -0.31 & 66.87 & -0.31 & 71.17 & 0.00 \\
electrical\_engineering & STEM & 55.86 & -0.69 & 57.24 & -0.34 & 55.52 & 0.00 \\
business\_ethics & Other & 57.00 & -0.50 & 63.50 & -0.50 & 62.50 & 0.00 \\
management & Other & 76.70 & 0.49 & 75.73 & -0.49 & 77.67 & 0.00 \\
high\_school\_government\_and\_politics & Social Sciences & 86.53 & -0.26 & 85.49 & 0.00 & 87.56 & 0.00 \\
professional\_psychology & Social Sciences & 63.24 & -0.08 & 63.89 & 0.33 & 64.05 & -0.16 \\
high\_school\_world\_history & Humanities & 78.06 & -0.42 & 77.64 & -0.00 & 79.11 & -0.21 \\
sociology & Social Sciences & 80.60 & 0.25 & 79.60 & 0.25 & 78.86 & -0.25 \\
human\_aging & Other & 67.26 & -0.45 & 66.59 & -0.22 & 67.26 & -0.45 \\
us\_foreign\_policy & Social Sciences & 82.50 & 0.00 & 80.50 & -0.50 & 80.00 & -0.50 \\
machine\_learning & STEM & 44.20 & -0.89 & 44.20 & -0.45 & 48.21 & -0.89 \\
public\_relations & Social Sciences & 62.73 & -1.36 & 60.00 & -1.82 & 64.55 & -0.91 \\
jurisprudence & Humanities & 74.54 & 0.46 & 73.61 & -0.46 & 72.22 & -0.93 \\
moral\_disputes & Humanities & 65.61 & -0.14 & 65.03 & -0.58 & 63.73 & -1.01 \\
\hline
\end{tabular}
\caption{\label{mmlu-gray-box-calibration}
Results of probability calibration method of GPT-3.5 model across 57 MMLU subtask.}
\end{table*}

\end{document}